\documentclass[sigconf]{acmart}

\usepackage{csquotes}
\usepackage{amsmath}
\usepackage{algorithm}
\usepackage[noend]{algpseudocode}
\usepackage{array}
\usepackage{caption}
\usepackage{subcaption}
\usepackage{ulem}
\usepackage{enumitem}

\AtBeginDocument{%
  \providecommand\BibTeX{{%
    \normalfont B\kern-0.5em{\scshape i\kern-0.25em b}\kern-0.8em\TeX}}}

\acmYear{2022}
\setcopyright{acmlicensed}\acmConference[CIKM '22]{Proceedings of the 31st
ACM International Conference on Information and Knowledge
Management}{October 17--21, 2022}{Atlanta, GA, USA}
\acmBooktitle{Proceedings of the 31st ACM International Conference on
Information and Knowledge Management (CIKM '22), October 17--21, 2022,
Atlanta, GA, USA}
\acmPrice{15.00}
\acmDOI{10.1145/3511808.3557322}
\acmISBN{978-1-4503-9236-5/22/10}

\begin{document}

\title{FedRN: Exploiting k-Reliable Neighbors Towards Robust Federated Learning}



\author{SangMook Kim}
\affiliation{%
 \institution{KAIST}
 \country{Republic of Korea}
 }
 \email{sangmook.kim@kaist.ac.kr}
 
\author{Wonyoung Shin}
\authornotemark[1]
\affiliation{%
 \institution{NAVER Shopping}
 \country{Republic of Korea}
 }
 \email{wonyoung.shin@navercorp.com}
 
 \author{Soohyuk Jang}
\affiliation{%
 \institution{POSTECH}
 \country{Republic of Korea}
 }
 \email{jang001@postech.ac.kr}
 \authornote{ Both authors contributed equally to this research.}
 
 \author{Hwanjun Song}
 \authornotemark[2]
\affiliation{%
 \institution{NAVER Corp.}
 \country{Republic of Korea}
 }
 \email{hwanjun.song@navercorp.com}

 \author{Se-Young Yun}
\affiliation{%
 \institution{KAIST}
 \country{Republic of Korea}
 }
 \email{yunseyoung@kaist.ac.kr}
 \authornote{ Both authors are corresponding authors of this paper.}


\begin{abstract}
Robustness is becoming another important challenge of federated learning in that the data collection process in each client is naturally accompanied by noisy labels. However, it is far more complex and challenging owing to varying levels of data heterogeneity and noise over clients, which exacerbates the client-to-client performance discrepancy. In this work, we propose a robust federated learning method called FedRN, which exploits $k$-reliable neighbors with high data expertise or similarity. Our method helps mitigate the gap between low- and high-performance clients by training only with a selected set of clean examples, identified by a collaborative model that is built based on the reliability score over clients. We demonstrate the superiority of \algname{} via extensive evaluations on three real-world or synthetic benchmark datasets. Compared with existing robust methods, the results show that \algname{} significantly improves the test accuracy in the presence of noisy labels. 
\end{abstract}

%
%

\settopmatter{printacmref=true}


\newcommand{\algname}{{FedRN}} 
\renewcommand{\shortauthors}{SangMook Kim et al.}

\maketitle


\section{Introduction}
\label{sec:introduction}

Federated learning\,(FL) is a privacy-preserving distributed learning technique, which handles a large corpus of decentralized data residing on multiple clients such as mobile or IoT devices\,\cite{bonawitz2019towards, yang2021characterizing}. The main idea is to learn a joint model by alternating local update and model aggregation phases. Each client performs the {local update} phase by training the current model with its own training data without sharing any of the clients' raw training data. Then, the {model aggregation} phase is executed by a server that merges received local models\,\cite{mcmahan2017communication}. As the need for on-device machine learning is rising, FL has attracted much attention in various fields\,\cite{yang2021flop, chen2021fedmatch, muhammad2020fedfast, jiang2019federated, luo2021fedskel, zhang2021desirable} and there have been numerous attempts to deploy it to on-device services, such as query suggestion for Google Keyboard\,\cite{yang2018applied} and question answering for Amazon Alexa\,\cite{chen2022self}.

\begin{figure}[t!]
\begin{center}
\includegraphics[width=8.4cm]{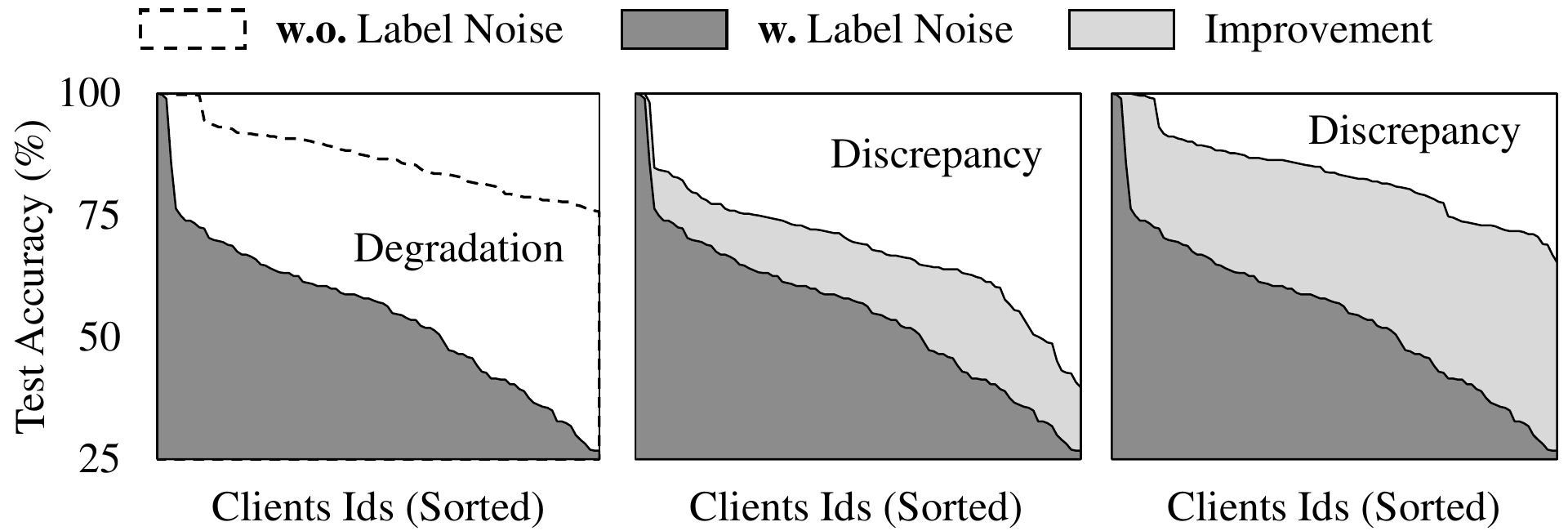}
\end{center}
\vspace*{-0.05cm}
\hspace*{0.65cm} {\small (a) FedAvg.} \hspace*{0.9cm} {\small (b) Co-teaching.} \hspace*{0.8cm} {\small (c) {\algname{}}.}
\vspace*{-0.3cm}
\caption{Performance difference of local models over 100 clients after training on CIFAR-10 with symmetric noise of $0-80\%$ in the Non-IID setting\,\cite{mcmahan2017communication}: (a) shows the performance degradation incurred by label noise in training data; (b) and (c) contrasts the improvement by Co-teaching and our proposed \algname{}. Client Ids are sorted in ascending order by the noise ratio of their local data. }
\label{fig:motivation}
\vspace*{-0.5cm}
\end{figure}


\begin{figure*}[t!]
\begin{center}
\includegraphics[width=16.0cm]{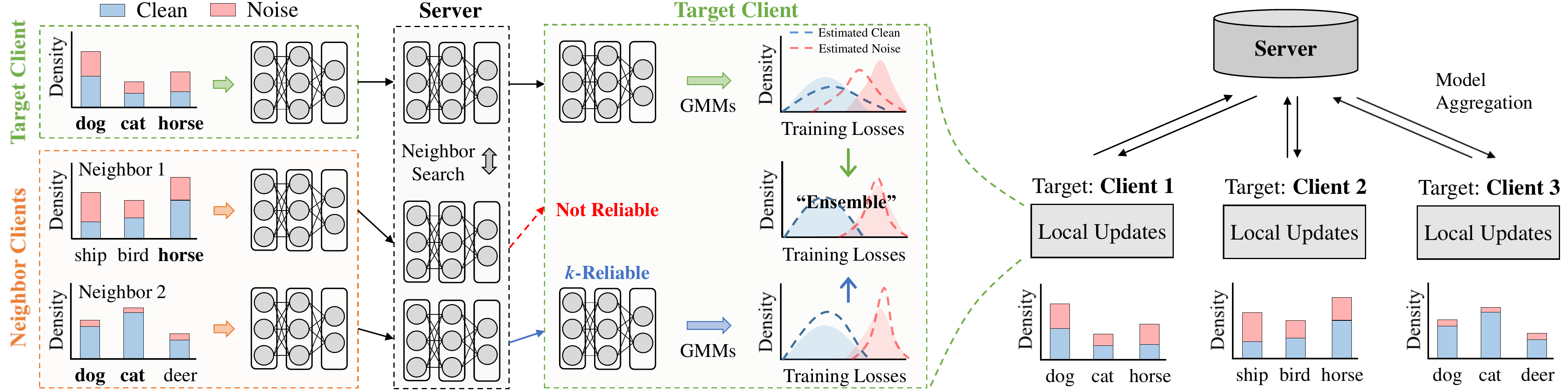}
\end{center}
\vspace*{-0.03cm}
\hspace*{3.0cm} {\small (a) Local Update Phase.} \hspace*{5.2cm} {\small (b) Model Aggregation Phase.} 
\vspace*{-0.3cm}
\caption{\textbf{Overview of \algname{} ($k=1$)}: (a) $k$-reliable neighbors are retrieved to fit GMMs to the loss values of all training examples in the target client. Their ensembled GMMs are leveraged to identify clean examples, which are used to update the target model in the local update phase; (b) The weights of updated local models are averaged in the model aggregation phase.}
\label{fig:key_idea}
\vspace*{-0.4cm}
\end{figure*}

Since the FL system is deployed over heterogeneous networks, \textit{robustness} has become an important challenge of federated learning, because the data collection process in each client is naturally accompanied by \textit{noisy labels}, which may be corrupted from the ground-truth labels\,\cite{lee2021machine}. In the presence of noisy labels, deep neural networks\,(DNNs) easily overfit to the noisy labels, thereby leading to poor generalization on unseen data\,\cite{zhang2016understanding}. Extensive efforts have been devoted to overcoming noisy labels in the centralized scenario\,\cite{ xia2019anchor, li2020dividemix}, but have yet to be studied widely in the federated learning\,(decentralized) scenario. {Handling noisy labels in the FL scenario is far more challenging than in the centralized scenario, particularly on the following \textit{two} difficulties: }
\begin{enumerate}[leftmargin=13pt]
\item \textbf{Data Heterogeneity}: Each client has non-independent and identically distributed\,(Non-IID) data with respect to the number of training examples for each class, which differ between clients.
\item \textbf{Varying Label Noise}: Noise ratios and types of data flaws vary between clients depending on how their data was collected, mainly attributed to the malfunction of data collectors, crowd-sourcing, and adversarial attacks.
\end{enumerate}

As shown in Figure \ref{fig:motivation}(a), these two difficulties exacerbate the performance discrepancy over clients because local data in each client varies in terms of {data heterogeneity} and {label noise}. Such a large performance discrepancy arguably results in poor performance of the global model, because client models are aggregated regardless of whether their models are corrupted from label noise\,\cite{chen2020fedbe}. 

The \textit{small-loss trick}, which updates a DNN with a certain number of small-loss examples every training iteration\,\cite{jiang2017mentornet, yu2019does, song2021robust, han2018co}, could be a prominent direction to provide robustness; small-loss examples are typically regarded as the clean set because DNNs tend to first learn from clean data and then gradually overfit to noisy data\,\cite{arpit2017closer}. However, as can be seen from Figure \ref{fig:motivation}(b), the performance discrepancy still remains even when all the local models were trained using a popular robust learning method, Co-teaching\,\cite{han2018co}, that uses the small-loss trick for sample selection. Although the overall performance is higher than that of FedAvg, the performance gap between the high- and low-performance model is considerably large. Therefore, it is essential to not only \textbf{(i)} identify clean examples from noisy data, but also \textbf{(ii)} {improve the low-performance models to increase the overall robustness for federated learning with noisy labels.} 


In this paper, we propose a simple yet effective robust approach called \textbf{\algname{}}\,({\underline{Fed}-erated learning with \underline{R}eliable \underline{N}eighbors}),
which exploits $k$-reliable neighbors
in the client pool {to help identify clean examples even when the target client is unreliable.}
We introduce the notion of \textit{k-reliable} neighbors, which are $k$ clients that have \textit{similar} data class distributions to the target client (data similarity), or \textit{minimal} label noise in their data (data expertise). Using $k$-reliable neighbors, \algname{} improves {the performance of every client's local model; the low-performance models benefit more from their $k$-reliable neighbors, as shown in Figure \ref{fig:motivation}(c).} 

%
In the local update phase shown in Figure \ref{fig:key_idea}(a), \algname{} first retrieves $k$-reliable neighbors from the server. \enquote{Neighbor 2} has a relatively clean dataset or a similar class distribution to the target client's, so the server transmits the model of \enquote{Neighbor 2} to the target client. The retrieved $k$-reliable models are used to identify clean examples from the client's dataset collaboratively with the target model; hence, with clear guidance from reliable neighbors, the target model can be improved significantly even if its  performance is poor due to heavy label noise.
%
In detail, the target and $k$-reliable models fit bi-modal univariate Gaussian mixture models\,(GMMs) to the loss distribution of the target's training data, respectively. Next, we build joint mixture models by aggregating all the GMMs based on the reliability scores of neighbor clients. 
Since the loss distributions of clean and noisy examples are bi-modal\,\cite{arpit2017closer}, the training examples with a higher probability of belonging to the clean\,(\textit{i.e.}, small-loss) modality are treated as clean examples and used to robustly update the target model. In this process, the low-performance model hardly hinders the accuracy of sample selection due to its low reliability score.

After the local update phase, the weights of updated models are averaged to build a global model in the model aggregation phase in Figure \ref{fig:key_idea}(b). 
These two phases are repeated until the global model converges following the standard federated learning pipeline\,\cite{mcmahan2017communication}. 

Our main contributions are summarized as follows:
\begin{itemize}[leftmargin=*]
\item This is a new FL framework that exploits reliable neighbors to tackle the challenge of data heterogeneity and varying label noise.
\item We introduce and examine the two indicators of data expertise and similarity, measuring the reliability score of neighbor clients without infringing on data privacy.
\item \algname{} remarkably {improves the performance of underperforming client models by mitigating the client-to-client discrepancy, thereby boosting overall robustness.}
\item \algname{} significantly outperforms state-of-the-art methods on three real-world or synthetic benchmark datasets with {varying levels} of data heterogeneity and label noise.
\end{itemize}
\section{Related Work}
\label{sec:related_work}

\noindent\textbf{Federated Learning.} 
Federated learning aims to learn a strong global model by exploiting all of the client data without infringing on data privacy. However, due to the heterogeneity in training data, it typically suffers from performance degradation of the global model\,\cite{zhu2021federated} or weight divergence\,\cite{zhao2018federated}.
In this regard, extensive efforts have been made to address these issues.
\citet{duan2019astraea} addressed the problem of long-tailed distribution in training data by performing data augmentation. \citet{zhao2018federated} let all of the clients share the same public data to mitigate the data heterogeneity. Meanwhile, \citet{li2018federated} and \citet{acar2020federated} added a regularization term that prevents the local model from diverging to the global model due to its own skewed Non-IID data. Although this family of methods contributes to improving the effectiveness of federated learning, they simply assume that all the labels in training data are clean, which is not a realistic scenario.

\smallskip
\noindent\textbf{Learning with Noisy Labels.} 
Extensive studies have been conducted to overcome noisy labels in the centralized scenario. 
A prominent direction for handling noisy labels is sample selection, which trains DNNs for a possibly clean subset of noisy training data.
For example, Co-teaching\,\cite{han2018co} trains two DNNs where each DNN selects small-loss examples in a mini-batch and then feeds them to its peer network for further training. MORPH\,\cite{song2021robust} divides the training process into two learning periods and employs two different criteria for sample selection. 
Another possible direction is to modify the loss of training examples based on importance reweighting or label correction\,\cite{patrini2017making, huang2020self, tanaka2018joint, zheng2021meta}. 

Most recently, to leverage all the training data, sample selection methods have been combined with other approaches, such as loss correction and semi-supervised learning. SELFIE\,\cite{song2019selfie} uses relabeled noisy examples in conjunction with the selected clean ones. DivideMix\,\cite{li2020dividemix} treats selected clean examples as labeled data and applies MixMatch\,\cite{berthelot2019mixmatch} for semi-supervised learning. Despite these methods' success, they do not perform well in the federated learning setting due to the client-to-client performance discrepancy. 

\smallskip
\noindent\textbf{Robust Federated Learning.} 
To address the noisy labels in FL, most studies assume that trustworthy data exists either on the client- or server-side. \citet{chen2020focus} estimated the credibility of each client with small clean validation data, and aggregated models based on the estimated credibilities of clients. \citet{tuor2021overcoming} proposed a data filtering method, which identifies highly relevant examples for the given specific task using a reference model trained on clean benchmark data. In another direction, a few robust methods using label correction have been proposed\,\cite{yang2022robust, zeng2022clc, xu2022fedcorr}. 
The most representative Robust FL\,\cite{yang2022robust} shares the central representations of clients' local data to maintain the consistent decision boundary over clients, and performed global-guided label correction assuming the IID scenario. However, existing methods often rely on unrealistic assumptions including the existence of clean validation data or the IID scenario. The recent label correction approaches including Robust FL still suffer from false label corrections produced by incorrect models under heavy label noise or severe Non-IID scenarios.

\smallskip
\noindent\textbf{Difference from Existing Work.} We clarify \textit{why FL with noisy labels is problematic}; the client-to-client performance discrepancy hinders the use of existing robust methods in the label noise community into FL. Although a few robust methods have been proposed, this problem has been overlooked.
In this paper, we thus bridge the two different topics to improve overall robustness {by introducing $k$-reliable neighbors such that they deliver credible information to the clients.} Moreover, compared to all aforementioned robust methods, \algname{} mainly focuses on identifying reliable neighbor clients for sample selection, which does not require any unrealistic supervision, such as a clean validation set or knowledge of true noise rates per client.

\section{Preliminaries}
\label{sec:preliminaries}

A multi-class classification problem requires training data $\mathcal{D}=\{(x_i, y^{*}_i)\}_{i=1}^{|\mathcal{D}|}$, where $x$ is a training example with its ground-truth label $y^{*}$. However, noisy training data $\tilde{\mathcal{D}}=\{(x_i, \tilde{y}_i)\}_{i=1}^{|\mathcal{D}|}$ may have corrupted labels such that $\tilde{y}_i \neq y^*_i$ for some $i$. We herein briefly summarize the conventional learning pipeline for {(i)} standard federated learning and {(ii)} robust learning with sample selection. \looseness=-1 

\begin{itemize}[leftmargin=*]
\item \textbf{Federated Learning\,(FedAvg):} It is assumed that each client ${\sf i}$ has its own Non-IID dataset $\mathcal{D}_{\sf i}$\,(i.e., $\mathcal{D} = \cup_{\sf i} \{\mathcal{D}_{\sf i}\}$), and cannot directly access another client's data. Therefore, a global model $\Theta_{{\sf global}}$ is trained by alternating the local update and model aggregation phase. 

During the local update phase, a fixed number of target clients $\mathcal{M}$ are selected from the entire client pool.
{For each target client ${\sf c} \! \in \! \mathcal{M}$, all the other clients are defined as its neighbors ${\sf n} \! \in \!\mathcal{N}_{{\sf c}}$.}
Next, the target client's local model $\Theta_{\sf c}$ is trained on its own data $\mathcal{D}_{\sf c}$ for certain local epochs with the standard stochastic gradient decent\,\cite{mcmahan2017communication}. All the trained models for $\mathcal{M}$ are then received and averaged by the server in the model aggregation phase,
\begin{equation}
\label{eq:aggregation}
\Theta_{{\sf global}} \leftarrow \sum_{{\sf c} \in \mathcal{M}} w_{\sf c} \Theta_{{\sf c}}, ~{\rm where}~ w_{\sf c}=\frac{|{\mathcal{D}}_{{\sf c}}|}{\sum_{{\sf c}^{\prime} \in \mathcal{M}}| {\mathcal{D}}_{{\sf c}'}|}
\end{equation}
where $w_{\sf c}$ is the aggregation weight that is based on the data size of each client. Lastly, the updated global model is broadcasted to the clients. This training round is called a transmission round and is repeated until convergence.
\smallskip
\item \textbf{Robust Learning with Sample Selection:} Many recent robust approaches have adopted sample selection, which treats a certain number of small-loss examples as the clean set\,\cite{han2018co, yu2019does, li2020dividemix, song2021robust}. In the centralized scenario, given a noisy mini-batch $\tilde{\mathcal{B}}$\,($\subset \tilde{\mathcal{D}}$), the clean subset $\mathcal{S}$ is identified and used to directly update the global model $\Theta_{\sf global}$,
\begin{equation}
\!\Theta_{\sf global}\! \leftarrow \Theta_{\sf global} - \eta\nabla\frac{1}{|\mathcal{S}|}\! \sum_{x \in \mathcal{S}} \!\ell(x, \tilde{y}; \Theta_{\sf global}),
\end{equation}
where $\eta$ is the learning rate and $\ell$ is a certain loss function. The remaining examples\,(mostly large-loss examples) in the mini-batch are excluded to pursue robust learning.  
\end{itemize}
Note that training data is typically assumed to be \textit{clean} in federated learning, while a global model is assumed to have access to \textit{all} of the training data in robust learning. However, our objective is to robustly train a global model $\Theta_{\sf global}$ even when noisy labels exist in the federated learning scenario.

\section{Proposed Method: FedRN}
\label{sec:methodology}

We introduce the notion of $k$-reliable neighbors and describe how to leverage them for robust federated learning with noisy labels.

\subsection{Main Concept: k-Reliable Neighbors}

The key idea of \algname{} is to exploit useful information in other clients without violating privacy. Hence, the main challenge is to find the most helpful neighbor clients using limited information about each client. Intuitively speaking, 
these helpful neighbor clients have (1)\,high expertise on their data, or (2)\,data distributions similar to the target client's: 

{
\renewcommand\labelenumi{(\theenumi)}
\begin{itemize}[leftmargin=*]
\item \textbf{Data Expertise:} The training data should be sufficiently \textit{clean}. The neighbor client with heavy noise could hinder identifying clean examples from the client's noisy data.

\item \textbf{Data Similarity:} Data and class distributions should be \textit{similar} to the target client's. The distribution shift problem could lead to model incompatibility between target and neighbor clients.  
\end{itemize}
}

We investigate how these two considerations affect selecting reliable neighbor clients for robust federated learning. Given a target client ${\sf c}$, a reliable neighbor client is defined to be one of the $k$-reliable neighbors $\mathcal{R}_{{\sf c}}(k)$, as described in Definition \ref{def:k_neighbor}. Satisfying the condition, $k$-reliable neighbors can deliver clean and useful information even when the training data of the target client is unreliable due to the heavy label noise.
%

\smallskip
\begin{definition} 
Let ${\rm Exp(\cdot)}$ and ${\rm Sim(\cdot, \cdot)}$ be the score function for data expertise of a client and data similarity between two clients, respectively. Given a target client ${\sf c}$ and available neighbor clients $\mathcal{N}_{{\sf c}}=\{{\sf n}_{i}\}_{i=1}^{|\mathcal{N}_{{\sf c}}|}$, the $k$-reliable neighbors $\mathcal{R}_{{\sf c}}(k)$ is a subset of $\mathcal{N}_{\sf c}$ with size $k$ satisfying, 
\begin{equation}
\begin{gathered}
\mathcal{R}_{{\sf c}}(k) = {\rm argmax}_{|(\mathcal{R}^{\prime} : {\sf n} \in \mathcal{R}^{\prime})| = k} ~ {\rm R}({\sf c}, {\sf n}),~{\rm where}\\
{\rm R}({\sf c}, {\sf n}) = \alpha \cdot {\rm Exp}({\sf n}) + (1-\alpha) \cdot {\rm Sim}({\sf c}, {\sf n})
\end{gathered}
\label{eq:reliability_score}
\end{equation}
and $\alpha$ is the balancing term that determines the contribution of each score. The data similarity to itself is 1, i.e., ${\rm Sim}({\sf c}, {\sf c})=1$. \qed
\label{def:k_neighbor}
\end{definition}

To compute the reliability score ${\rm R}({\sf c}, {\sf n})$ in Eq.\,\eqref{eq:reliability_score}, we propose two indicators of data expertise and similarity using client's local models. Since it is infeasible to access the raw data in neighbor clients, we leverage the server's copies of local models, which are sent to the server during the model aggregation phase.
By doing so, our approach does not compromise the privacy-preserving property in federated learning. 

With the intention of selecting clients with cleaner examples, we define the data expertise score motivated by the memorization effect of DNNs.
In general, DNNs memorize clean examples faster than noise examples during training, which has been observed consistently even at extreme noise ratios\,\cite{song2021robust}.
In this sense, the early memorization of clean examples leads to higher training accuracy if the client has cleaner local data. 
Consequently, during the model aggregation phase, the server receives the training accuracy of each client along with their local models, then computes and normalizes the data expertise Exp($\cdot$) per client,\looseness=-1
\begin{equation}
\!{\rm Exp}({\sf c}) = \frac{{\rm Acc}({\sf c}) - {\rm minAcc}(\{{\sf c}\} \cup \mathcal{N}_{\sf c})}{{\rm maxAcc}(\{{\sf c}\} \cup \mathcal{N}_{\sf c}) - {\rm minAcc}(\{{\sf c}\} \cup \mathcal{N}_{\sf c})},\!\!
\label{eq:expertise}
\end{equation}
where ${\rm Acc}$ denotes the training accuracy of the local model for the client ${\sf c}$; ${\rm maxAcc}$ and ${\rm minAcc}$ are the maximum and minimum training accuracy of the given client set.

Regarding the data similarity score, we propose to use \textit{predictive difference}, which provides a hint to the similarity between two heterogeneous data distributions {without} the use of raw data. Hence, the data similarity between two clients is approximated by the prediction difference for the \textit{same} random input, where a smaller difference indicates a higher similarity. 
Accordingly, after the local update phase, each client generates the same input $\tilde{x}$ of Gaussian random noise\,\cite{jeong2020federated} and transmits its softmax outputs $p(\tilde{x};\Theta_{{\sf n}})$ into the server, where $p(x;\Theta_{{\sf n}})$ denotes the softmax output of the client ${\sf n}$ for input $x$. The data similarity Sim($\cdot, \cdot$) between the target client ${\sf c}$ and neighbor client ${\sf n}$ is then computed by the server, 
\begin{equation}
{\rm Sim}({\sf c}, {\sf n}) = {\rm Cosine}\big(p(\tilde{x};\Theta_{{\sf c}}), p(\tilde{x};\Theta_{{\sf n}})\big).
\label{eq:similarity}
\end{equation}
We use the cosine similarity denoted by Cosine to measure the predictive difference, because the \textit{symmetry} of similarity is guaranteed, i.e., ${\rm Sim}({\sf c}, {\sf n})={\rm Sim}({\sf n}, {\sf c})$. The min-max normalization is also applied similarly to data expertise. We provide an in-depth analysis of how the two proposed indicators work precisely in Section \ref{sec:in_depth_metric}. 

\subsection{Robust Learning with k-Reliable Neighbors}

For every communication round,  each client participating in the upcoming round receives its $k$-reliable neighbors from the server. 
When updating the target client's model, we follow the general philosophy of \textit{sample selection} for robust learning by identifying clean examples from the noisy training data. 

Different to robust methods in centralized scenarios, $k+1$ models, comprising $k$-reliable neighbors and the target client's model, cooperate to mitigate the performance discrepancy among clients and obtain more reliable results.
For each of the $k+1$ models, we fit two-component GMMs to model the loss distributions of clean and noisy examples in view of loss distributions being \textit{bi-modal}\,\cite{li2020dividemix, arazo2019unsupervised}. Based on the ensembled results of these GMMs, the target client's model is only updated with examples selected as clean.  

To be specific, our aim is to estimate the probability of each training example being clean with $k+1$ models with different reliability scores. At the beginning of the local update phase, the GMM is fitted to the loss of all available training examples in the target client by using the Expectation-Maximization\,(EM) algorithm. Given a noisy example $x$, the probability of being clean is obtained through its posterior probability for the clean\,(small-loss) modality,
\begin{equation}
p \big( g|\ell(x, \tilde{y};\Theta) \big) = p(g) p\big(\ell(x, \tilde{y}; \Theta) | g \big) / p\big(\ell(x, \tilde{y};\Theta)\big),
\end{equation}
where g denotes the Gaussian modality with a smaller mean\,(i.e., smaller loss) and $\ell(\cdot)$ is the loss function. Subsequently, the probability of each example to be clean is ensembled over $k+1$ models with their reliability scores ${\rm R}(\cdot, \cdot)$ in Eq.\,\eqref{eq:reliability_score}. Given a target client ${\sf c}$, the clean probability of joint mixtures can be formulated as:
\begin{equation}
\begin{gathered}
p\big( {\rm clean} | x ; \mathcal{R}_{{\sf c}}(k) \big)= \!\!\!\!\!\!\!\!\!\sum_{{\sf n} \in \{{\sf c}\} \cup \mathcal{R}_{{\sf c}}(k) }\!\!\!\!\!\!\!\! {\rm R}^{\prime}({\sf c}, {\sf n}) \times p \big( g|\ell(x, \tilde{y} ;\Theta_{{\sf n}}) \big), \!\!\! \\
{\rm where} ~~ {\rm R}^{\prime}({\sf c}, {\sf n}) = {\rm R}({\sf c}, {\sf n}) ~~~ / \!\!\!\!\!\!\!\!\sum_{{\sf n}^{\prime} \in \{ {\sf c}\} \cup \mathcal{R}_{{\sf c}}(k)}\!\!\!\!\! {\rm R}({\sf c}, {\sf n}^{\prime}).\!\!\!
\end{gathered}
\end{equation}
The ensemble with the reliability scores give higher weight to the GMMs of trusted neighbors with high data expertise or similarity.

Lastly, \algname{} constructs a clean set $\mathcal{S}_{{\sf c}}$ such that each training example in the set has a clean probability greater than $0.5$, 
\begin{equation}
\mathcal{S}_{{\sf c}} = \{ x \in \tilde{\mathcal{D}}_{{\sf c}} : p({\rm clean} | x ; \mathcal{R}_{{\sf c}}(k)) > 0.5 \},
\label{eq:selection}
\end{equation}
where $\tilde{\mathcal{D}}_{{\sf c}}$ is the noisy data of the target client ${\sf c}$. The model is finally updated with only the clean set for a specified number of local epochs in the local update phase, while the complement of the set, which are likely to be noisy, are discarded for robust learning. 

This collaborated approach with $k$-reliable neighbors considerably increases the robustness of \textit{low}-performance clients 
with the aid of neighbors with high data expertise or similarity. 
Therefore, \algname{} reduces the performance gap between clients and enhances the overall robustness, thus leading to a stronger global model. 

\subsection{Fine-tuning with k-Reliable Neighbors}
Due to the data heterogeneity in federated learning, the problem of distribution shift between clients' local models is another challenge when ensembling their predictions for sample selection. 
Hence, \algname{} is integrated with the fine-tuning technique that helps quickly adapt to the local data distribution\,\cite{collins2021exploiting}.

Before constructing the clean set in Eq.\,\eqref{eq:selection}, we identify an auxiliary set of clean examples using \textit{only} the target client's model,
\begin{equation}
\mathcal{S}_{{\sf c}}^{\sf aux} = \{ x \in \tilde{\mathcal{D}}_{{\sf c}} : p({\rm clean} | x ; \{{\sf c}\}) > 0.5 \}
\label{eq:aux_set}
\end{equation}
\vspace{-0.1cm}
and fine-tune all the retrieved $k$-reliable neighbors for the auxiliary set. The auxiliary set could, however, be very noisy especially when the target client has heavy noise in its training data. To prevent feature extractors from being corrupted by such client-side noise, we only fine-tune the classification head. In Section \ref{sec:analysis_fine_tune}, we examine the effect of the fine-tuning technique on data expertise and similarity used in neighbor search, and provide insights on their use for robust federated learning with noisy labels.

\vspace*{-0.1cm}
\subsection{Algorithm Pseudocode}
\vspace*{-0.0cm}

\setlength{\textfloatsep}{8pt}
\begin{algorithm}[t!]
\caption{FedRN Algorithm}
\label{alg:fedrn}
\hspace*{-0.8cm}\textsc{Input:} $\tilde{\mathcal{D}}_{{\sf c}}$: noisy data of client ${\sf c}$, $rounds$: $\#$ total rounds,\\ $e$: $\#$ local epochs, $k$: $\#$ reliable neighbors 

\hspace*{-3.9cm}\textsc{Output:} $\Theta_{{\sf global}}^{t}$: a global model 
\begin{algorithmic}[1]
\State $t \leftarrow 1$, ~$\Theta_{{\sf global}}^t \leftarrow \text{Initialize network parameters}$
\For{$round=1$ {\bf to} $rounds$}
\State $\mathcal{M} \leftarrow \text{Randomly select $m$ clients}\,(m > k)$
\State /* Find $k$-reliable neighbors in Definition \ref{def:k_neighbor} */
\State Compute $\forall_{{\sf c} \in \mathcal{M}}\forall_{{\sf n} \in \mathcal{N}_{\sf c}} ~{\rm Sim}({\sf c}, {\sf n})$ by Eq.\,\eqref{eq:similarity}
\State Retrieve $\mathcal{R}_{{\sf c}}(k)$ for ${\sf c} \in \mathcal{M}$
\State Broadcast $\Theta_{{\sf global}}^t$ and $\mathcal{R}_{{\sf c}}(k)$ to clients $\in \mathcal{M}$
\State {\color{blue} /*  {I. Local Update Phase} */} 
\For{${\sf c} \in \mathcal{M}$ in parallel}
\State $\mathcal{R}_{{\sf c}}(k) \leftarrow \text{Receive $k$-reliable neighbors}$
\State $\mathcal{R}_{{\sf c}}(k) \leftarrow $ Fine-tune for $\mathcal{S}_{\sf c}^{\sf aux}$ in Eq.\,\eqref{eq:aux_set}.
\State /* Sample selection with $\mathcal{R}_{{\sf c}}(k)$ by Eq.\,\eqref{eq:selection} */
\State $\mathcal{S}_{{\sf c}} \leftarrow \{ x \in \tilde{\mathcal{D}}_{{\sf c}} : p({\rm clean} | x ; \mathcal{R}_{{\sf c}}(k)) > 0.5 \}$ 
\State $\Theta_{{\sf c}} \leftarrow \Theta_{{\sf global}}^{t}$ /* Initialization */ 
\State $\Theta_{{\sf c}} \leftarrow $ Train for $e$ epochs with $\mathcal{S}_{{\sf c}}$
\State ${\rm Acc}({\sf c}) \leftarrow $ Compute data expertise by Eq.\,\eqref{eq:expertise}
\State $p(\tilde{x}; \Theta_{\sf c}) \leftarrow$ Compute the softmax output for $\tilde{x}$
\State Send $\Theta_{{\sf c}}$, ${\rm Acc}({\sf c})$, and $p(\tilde{x}; \Theta_{\sf c})$ to the server
\EndFor
\State {\color{blue} /*  {II. Model Aggregation Phase} */} 
\State $\Theta_{{\sf global}}^{t+1} \leftarrow \sum_{{\sf c} \in \mathcal{M}} w_{\sf c} \Theta_{{\sf c}}$ s.t. $w_{\sf c}=\frac{|\tilde{\mathcal{D}}_{{\sf c}}|}{\sum_{{\sf c}^{\prime} \in \mathcal{M}}| \tilde{\mathcal{D}}_{{\sf c}'}|}$ 
\State $t \leftarrow t + 1$
\EndFor
\State \Return {$\Theta_{{\sf global}}^t$}
\end{algorithmic}
\end{algorithm}

Algorithm \ref{alg:fedrn} describes the overall procedure of \algname{}.
We perform standard federated learning, FedAvg, for {warm-up} epochs before applying \algname{}, because robust training methods are generally applied after the warm-up phase in the literature\,\cite{han2018co, song2019selfie}. 
When training with \algname{}, retrieved neighbor models are fine-tuned for 1 epoch before sample selection\,(Line 11). 
{Next, the target client identifies clean examples by leveraging $k$-reliable neighbor models\,(Line 13). The received global model is updated with estimated clean samples\,(Lines 14--15). In the aggregation phase, the central server aggregates the updated local models\,(Line 20).
}

The main additional cost of \algname{} is the communication overhead from sending $k$-reliable models from the server to the selected client\,(see Appendix \ref{sec:cost_analysis} for detailed cost analysis). However, a small number of reliable neighbors is sufficient (see Section \ref{sec:aba_neighbor}) and the cost can be drastically reduced by over a factor of 32 times in modern federated learning by sending the difference of parameters\,\cite{jeong2020federated} or a compressed model\,\cite{malekijoo2021fedzip}. Furthermore, there is no issues in model convergence because we applied a sample selection approach, which don't has convergence problems, in a decentralized setting using FedAvg\,\cite{mcmahan2017communication}, which has proven model convergence.

\section{Evaluation}
\label{sec:evaluation}

Our evaluation was conducted to support the following:
\vspace*{-0.05cm}
\begin{itemize}[leftmargin=*]
\item \algname{} is \textbf{more robust} than {five} state-of-the-art methods for federated learning with noisy labels\,(Section \ref{sec:robust_comparison}).
\vspace*{0.05cm}
\item \algname{} consistently identifies clean examples from noisy data with \textbf{high precision} and \textbf{recall}\,(Section \ref{sec:selection_comparison}).
\vspace*{0.05cm}
\item The reliability metric is \textbf{effective} in finding $k$-reliable neighbors for robust learning\,(Section \ref{sec:in_depth_metric}).
\vspace*{0.05cm}
\item The use of fine-tuning is \textbf{necessary} owing to the distribution shift in local data\,(Section \ref{sec:analysis_fine_tune}).
\vspace*{0.05cm}
\item A small number\,($k\leq 2)$ of reliable neighbors is \textbf{sufficient} to achieve high robustness\,(Section \ref{sec:aba_neighbor}).
\end{itemize}
\vspace*{-0.12cm}

\subsection{Experiment Configuration}

We verified the superiority of \algname{} compared with five robust learning methods on three real-world or synthetic benchmark datasets in a Non-IID federated learning setting, where the Non-IID setting is more realistic and difficult than the IID setting\,\cite{zhu2021federated, zhao2018federated, briggs2020federated}.

\smallskip
\noindent\textbf{Datasets.} We performed image classification on synthetic and real-world benchmark data: CIFAR-10 and CIFAR-100; mini-WebVision, a subset of real-world noisy data consisting of large-scale web images. We used the first 50 classes in WebVision V1 following the literature\,\cite{li2020dividemix}. The statistics of them are summarized in Table \ref{tab:summary_data}.


{
\newcolumntype{L}[1]{>{\raggedright\let\newline\\\arraybackslash\hspace{0pt}}m{#1}}
\newcolumntype{X}[1]{>{\centering\let\newline\\\arraybackslash\hspace{0pt}}p{#1}}
\begin{table}[t!]
\centering
\resizebox{\linewidth}{!}{%
\begin{tabular}{c|cccc}
\toprule
& \# of Train & \# of Val.  & \# of Classes   & Noise Ratio        \\ \midrule
CIFAR-10  & 50,000          & 10,000            & 10     & $\approx 0\%$          \\ 
CIFAR-100 & 50,000          & 10,000            & 100     & $\approx 0\%$                 \\ 
mini-WebVision & 65,944          & 2500              & 50       & $\approx 20\%$                 \\ \bottomrule
\end{tabular}%
}
\vspace*{-0.05cm}
\caption{Summary of datasets.}
\label{tab:summary_data}
\vspace*{-0.55cm}
\end{table}
}

{
\newcolumntype{L}[1]{>{\raggedright\let\newline\\\arraybackslash\hspace{0pt}}m{#1}}
\newcolumntype{X}[1]{>{\centering\let\newline\\\arraybackslash\hspace{0pt}}p{#1}}
\begin{table*}[!t]
\small
\begin{tabular}{L{2.2cm} |X{0.91cm}X{0.91cm}|X{0.91cm}|X{0.91cm}|X{0.91cm}X{0.91cm}|X{0.91cm}|X{0.91cm}|X{0.91cm}X{0.91cm}|X{0.91cm}|X{0.91cm}}\toprule
Non-IID Type &  \multicolumn{4}{c|}{Shard\,($S=2$)} & \multicolumn{4}{c|}{Shard\,($S=5$)} & \multicolumn{4}{c}{Dirichlet\,($\beta=0.5$)}  \\\midrule
Noise Type & \multicolumn{2}{c|}{Symmetric} & {Asym} & {Mixed} & \multicolumn{2}{c|}{Symmetric} & {Asym} & {Mixed} & \multicolumn{2}{c|}{Symmetric} & {Asym} & {Mixed} \\
Noise Rate & \!\!0.0--0.4\!\! & \!\!0.0--0.8\!\! & \!\!0.0--0.4\!\! & \!\!0.0--0.4\!\! & \!\!0.0--0.4\!\! & \!\!0.0--0.8\!\! & \!\!0.0--0.4\!\! & \!\!0.0--0.4\!\! & \!\!0.0--0.4\!\! & \!\!0.0--0.8\!\! & \!\!0.0--0.4\!\! & \!\!0.0--0.4\!\! \\\toprule
Oracle\footnote{}     & 69.10 & 67.19 & 69.10 & 68.89 & 78.00 & 75.94 & 78.00 & 77.7 & 81.79 & 80.38 & 81.79 & 81.39 \\ \midrule
FedAvg\,\cite{mcmahan2017communication}     & 46.39 & 38.69 & 49.09 & 46.82 & 66.44 & 56.46 & 67.23 & 67.95 & 75.92 & 70.60 & 77.77 & 77.17 \\
Co-teaching\,\cite{han2018co}\!\! & 63.20 & 52.72 & 62.48 & 61.68 & 74.66 & 67.17 & 74.18 & 75.18 & 79.97 & 75.39 & 79.53 & 79.94 \\
Joint-optm\,\cite{tanaka2018joint}  & 50.44 & 42.23 & 38.04 & 47.28 & 69.22 & 64.02 & 67.29 & 68.84 & 75.46 & 70.43 & 74.81 & 75.43 \\
SELFIE\,\cite{song2019selfie}      & 62.64 & 53.69 & 64.21 & 62.63 & 74.57 & 66.90 & 74.77 & 74.44 & 78.57 & 72.92 & 78.58 & 79.02 \\
DivideMix\,\cite{li2020dividemix}   & 62.35 & 58.18 & 62.07 & 63.38 & 68.73 & 65.82 & 68.95 & 69.32 & 74.26 & 72.25 & 73.29 & 73.47 \\  
Robust FL\,\cite{yang2022robust} & 56.25 & 45.59 & 55.52 & 57.58 & 70.30 & 62.89 & 69.04 & 70.02 & 75.75 & 70.63 & 74.00 & 75.49 \\
\midrule
\textbf{\algname{} (k=1)}  & 67.33 & 60.33 & 67.51 & 67.92 & 76.37 & 72.30 & 76.74 & 76.92 & 79.99 & 75.92 & 80.05 & 79.79 \\
\textbf{\algname{} (k=2)} & \textbf{67.62} & \textbf{62.94} &  \textbf{68.33} & \textbf{68.11} & \textbf{76.81} & \textbf{72.85} & \textbf{77.33} & \textbf{76.99} & \textbf{80.34} & \textbf{76.49} & \textbf{80.38} & \textbf{80.28} \\
\bottomrule
\end{tabular}
\vspace*{-0.1cm}
\caption{Test accuracy\,(\%) on CIFAR-10 with symmetric, asymmetric\,(Asym), and mixed noise\,(Mixed).}
\label{tab:cifar10}
\vspace*{-0.75cm}
\end{table*}
}


{
\newcolumntype{L}[1]{>{\raggedright\let\newline\\\arraybackslash\hspace{0pt}}m{#1}}
\newcolumntype{X}[1]{>{\centering\let\newline\\\arraybackslash\hspace{0pt}}p{#1}}
\begin{table}[!t]
\small
\begin{tabular}{L{2.2cm} |X{1.14cm}X{1.14cm}|X{1.14cm} |X{1.14cm}}\toprule
Non-IID Type &  \multicolumn{4}{c}{Shard\,($S=20$)} \\\midrule
Noise Type & \multicolumn{2}{c|}{Symmetric} & {Asym} & {Mixed} \\
Noise Rate & \!\!0.0--0.4\!\! & \!\!0.0--0.8\!\! & \!\!0.0--0.4\!\! & \!\!0.0--0.4\!\! \\\toprule
Oracle      & 47.83 & 44.82 & 47.83 & 47.33 \\\midrule
FedAvg\,\cite{mcmahan2017communication}      & 33.36 & 26.15 & 34.90 & 34.02 \\
Co-teaching\,\cite{han2018co} & 43.43 & 37.05 & 42.90 & 43.83 \\
Joint-optm\,\cite{tanaka2018joint}   & 35.89 & 30.92 & 33.15 & 35.29 \\
SELFIE\,\cite{song2019selfie}      & 44.14 & 37.65 & 43.25 & 43.78 \\
DivideMix\,\cite{li2020dividemix}   & 46.21 & 42.83 & 45.59 & 46.94 \\ 
Robust FL\,\cite{yang2022robust}   & 35.33 & 29.20 & 33.08 & 34.27 \\\midrule
\textbf{\algname{} (k=1)} & 47.30 & 42.27 & \textbf{47.62} & 46.62 \\
\textbf{\algname{} (k=2)} & \textbf{47.46} & \textbf{43.07} & 47.52 & \textbf{47.17} \\
\bottomrule
\end{tabular}
\vspace*{-0.1cm}
\caption{Test accuracy\,(\%) on CIFAR-100.}
\label{tab:cifar100}
\vspace*{-0.5cm}
\end{table}
}

For the federated learning setup with noisy labels, we merged the standard setting in the two research communities:
\begin{itemize}[leftmargin=*]
\item \underline{Non-IID Data:} Two popular ways of client data partitioning were applied\,\cite{zhu2021federated}; \textbf{(i)}\,Sharding: training data is sorted by its class labels and divided into {\small \enquote{$S \!  \times \! (\#~{\rm clients})$}} shards, which are randomly assigned to each client with an equal number of $S$. \textbf{(ii)}\,Dirichlet Distribution: each client is assigned with training data drawn from the Dirichlet distribution with a concentration parameter $\beta$. For the experiments, $S$ was set to  $2$, $5$, $10$, or $20$, and $\beta$ was set to $0.5$. \looseness=-1
\vspace*{0.05cm}
\item \underline{Label Noise:} We injected artificial noise by following typical protocols\,\cite{song2020learning}; \textbf{(i)}\,Symmetric Noise: a true label is flipped into all possible labels with equal probability. \textbf{(ii)}\,Asymmetric Noise: a true label is flipped into only a certain label. Note that we injected \textbf{varying levels} of label noise into the clients, so each client had different noise rates. For instance, a noise of $0.0$--$0.8$ means that the noise rate increases linearly from $0.0$ to $0.8$ as the client's index increases; thus, the average noise rate of clients' training data is $0.4$. We further tested for \textbf{mixed noise} where the two different types of label noise are injected into the clients. That is, symmetric noise is applied to half of the clients, while asymmetric noise is applied to the remaining clients.
\end{itemize} 
\vspace*{-0.0cm}
We did not inject any label noise into mini-WebVision because it contains real label noise whose rate is estimated at $20.0\%$\,\cite{song2020learning}.

\smallskip
\noindent\textbf{Implementation Details.} 
We compared \algname{} with a standard federated learning method, FedAvg\,\cite{mcmahan2017communication}, and {five} state-of-the-art robust methods for handling noisy labels, namely Co-teaching \cite{han2018co}, Joint-optimization \cite{tanaka2018joint}, SELFIE \cite{song2019selfie}, DivideMix \cite{li2020dividemix}, {and Robust FL\,\cite{yang2022robust}}. All the robust methods were combined with FedAvg to support decentralized learning. The total number of clients was set to $100$ with a fixed participation rate of $0.1$. The training\,(transmission) round was set to be $500$ for CIFAR-10 and $1000$ for CIFAR-100 and WebVision, ensuring all the models' training convergence.
The SGD optimizer was used with a learning rate of $0.01$ and a momentum of $0.5$. The number of local epochs was set to be $5$. 
We used the model for learning consists of 4 convolution layers and 1 linear classifier for CIFAR-10 and MobileNet for CIFAR-100 and mini-WebVision. 

\footnotetext[1]{Oracle trains the network using FedAvg with all the true-labeled examples, i.e., $(1-\text{noise rate}) \times 100 \%$ of all the examples.}

{As for hyperparameters related to our method, \algname{} requires the balancing term $\alpha$ and number of reliable neighbors $k$. We used $\alpha=0.6$, which was obtained via a grid search\,(see the section \ref{sec:parameter_search}). We used $k=1$ and $k=2$ since the performance gain of \algname{} was consistently high as long as $k\geq 1$, which is detailed in Section \ref{sec:aba_neighbor}.} 

The hyperparameters of the baseline robust methods were configured favorably in line with the literature. We clarify the hyperparmeter setup of all baseline algorithms, as follows:
\begin{itemize}[leftmargin=*]
\item Joint Optimization\,\cite{tanaka2018joint} requires the two coefficients for its two regularization losses for robust learning. These two coefficients were set to be 1.2 and 0.8, respectively.
\smallskip
\item SELFIE\,\cite{song2019selfie} requires the uncertainty threshold and the length of label history as its hyperparameters. As suggested by the authors, they were set to be 0.05 and 15, respectively. 
\smallskip
\item DivideMix\,\cite{li2020dividemix} applies MixMatch for semi-supervised learning, which requires the sharpening temperature, the unsupervised loss coefficient, and the Beta distribution parameter. They were set to be 0.5, 25, and 4, respectively.
\smallskip
\item Robust FL\,\cite{yang2022robust} requires two coefficients for its two additional regularization losses similar to Joint Optimization. They were set to be 1.0 and 0.8, respectively.
\end{itemize}

 


{
\newcolumntype{L}[1]{>{\raggedright\let\newline\\\arraybackslash\hspace{0pt}}m{#1}}
\newcolumntype{X}[1]{>{\centering\let\newline\\\arraybackslash\hspace{0pt}}p{#1}}
\begin{table}[!t]
\small
\begin{tabular}{L{2.2cm} |X{2.65cm}|X{2.65cm}}\toprule
Shard ($S=10$) & \multicolumn{2}{c}{mini-WebVision}\\
Method      & Top-1 Accuracy & Top-5 Accuracy \\\midrule
FedAvg\,\cite{mcmahan2017communication}      & 20.80 & 45.00 \\
Co-teaching\,\cite{han2018co} & 21.76 ($+$\,0.96) & 46.64 ($+$\,1.64) \\
Joint-optm\,\cite{tanaka2018joint}   & 13.56 ($-$\,7.24) & 35.56 ($-$\,9.44) \\
SELFIE\,\cite{song2019selfie}      & 22.20 ($+$\,1.40) & 48.76 ($+$\,3.76) \\
DivideMix\,\cite{li2020dividemix}   & 20.84 ($+$\,0.04) & 45.72 ($+$\,0.72)  \\ 
Robust FL\,\cite{yang2022robust}   & 14.12 ($-$\,6.68) & 35.24 ($-$\,9.76) \\
\midrule
\textbf{\algname{} (k=1)}  & 22.52 ($+$\,1.72) & 48.48 ($+$\,3.48) \\
\textbf{\algname{} (k=2)}  & \textbf{22.76 ($+$\,1.96)} & \textbf{49.16 ($+$\,4.16)} \\\bottomrule
\end{tabular}
\vspace*{0.1cm}
\caption{Validation accuracy\,(\%) on mini-WebVision. The values in parentheses are the improvement over FedAvg.}
\label{tab:webvision}
\vspace*{-0.7cm}
\end{table}
}


All of the algorithms were implemented using PyTorch $1.7.0$ and executed using four NVIDIA RTX 2080Ti GPUs. 
In support of reliable evaluation, we repeated every task thrice and reported the average test\,(or validation) accuracy, which is the common measure of robustness to noisy labels\,\cite{han2018co, chen2019understanding}. 

\begin{figure*}[!ht]
\centering
\includegraphics[width=\linewidth]{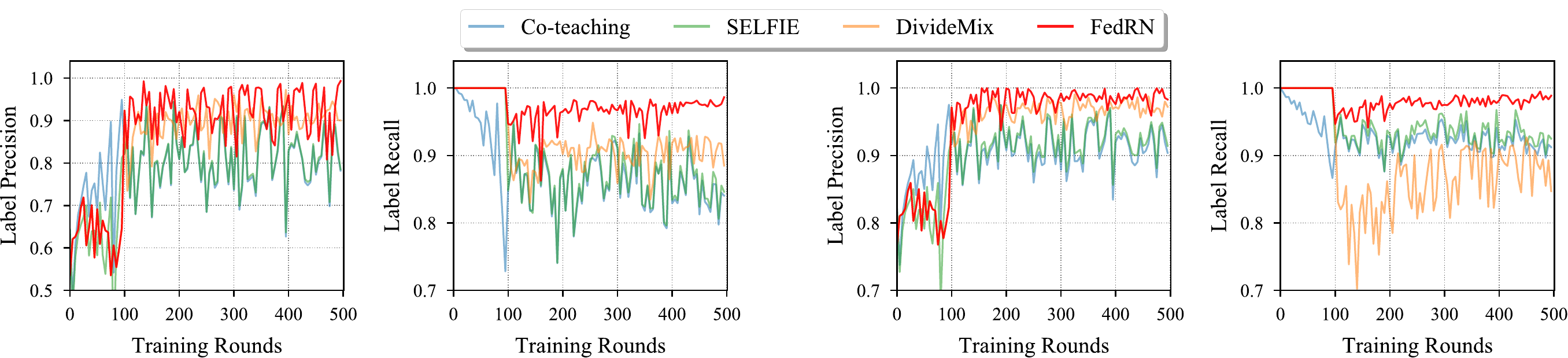}
\begin{minipage}[t]{.45\linewidth}
\centering
\label{label-sym80}
\end{minipage}%
\hskip 1cm
\begin{minipage}[t]{.45\linewidth}
\centering
\label{label-asym40}
\end{minipage}
\vspace*{0.05cm}
\hspace*{-0.4cm} {\small (a) Symmetric Noise of $0.0$--$0.8$.} \hspace*{5.6cm} {\small (b) Asymmetric Noise of $0.0$--$0.4$.} 
\vspace*{-0.35cm}
\caption{Label precision and label recall curves on CIFAR-10 with the shard\,($S=2$) setting.}
\label{fig:label-pr}
\vspace*{-0.25cm}
\end{figure*}

\begin{figure}
\centering
\begin{subfigure}[b]{0.45\linewidth}
    \centering
    \includegraphics[width=\textwidth]{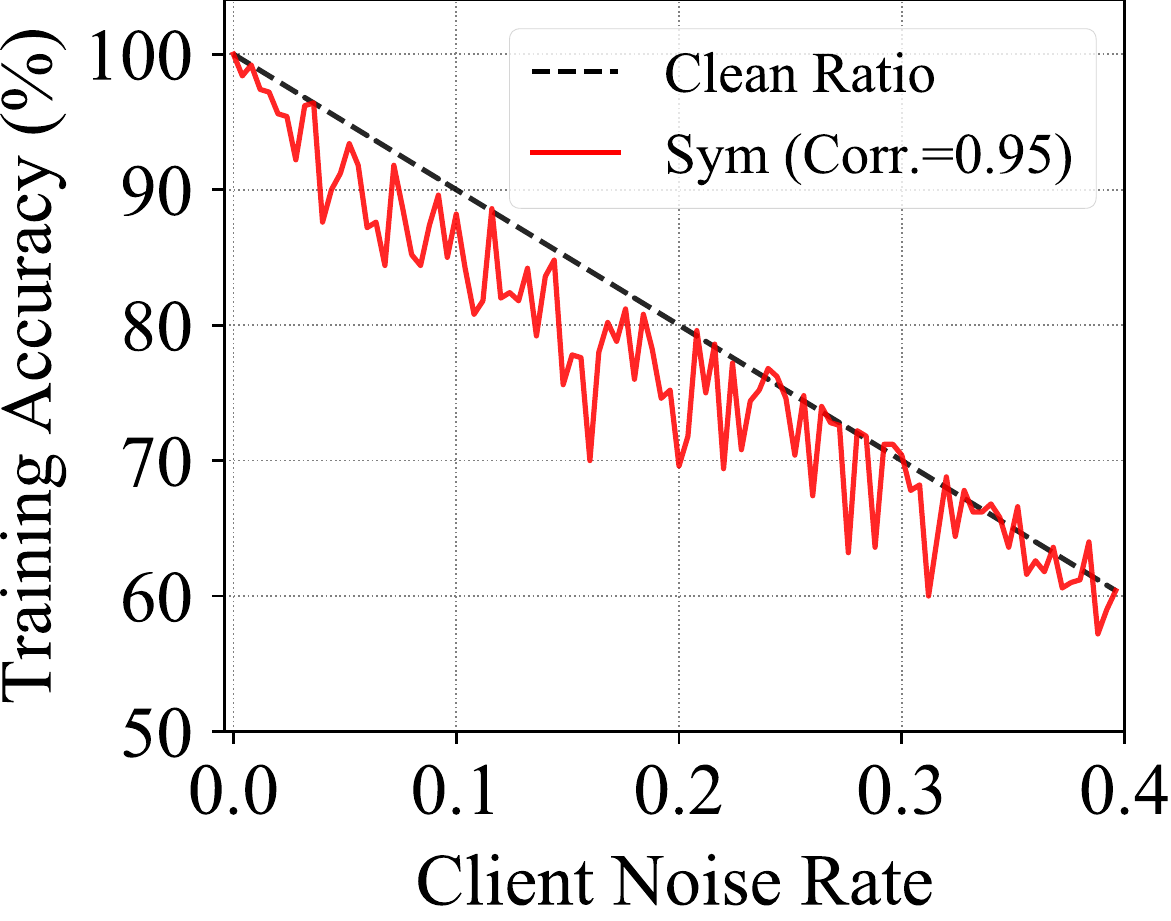}
    \label{fig:expertise-sym}
\end{subfigure}
\hfill
\begin{subfigure}[b]{0.45\linewidth}
    \centering
    \includegraphics[width=\textwidth]{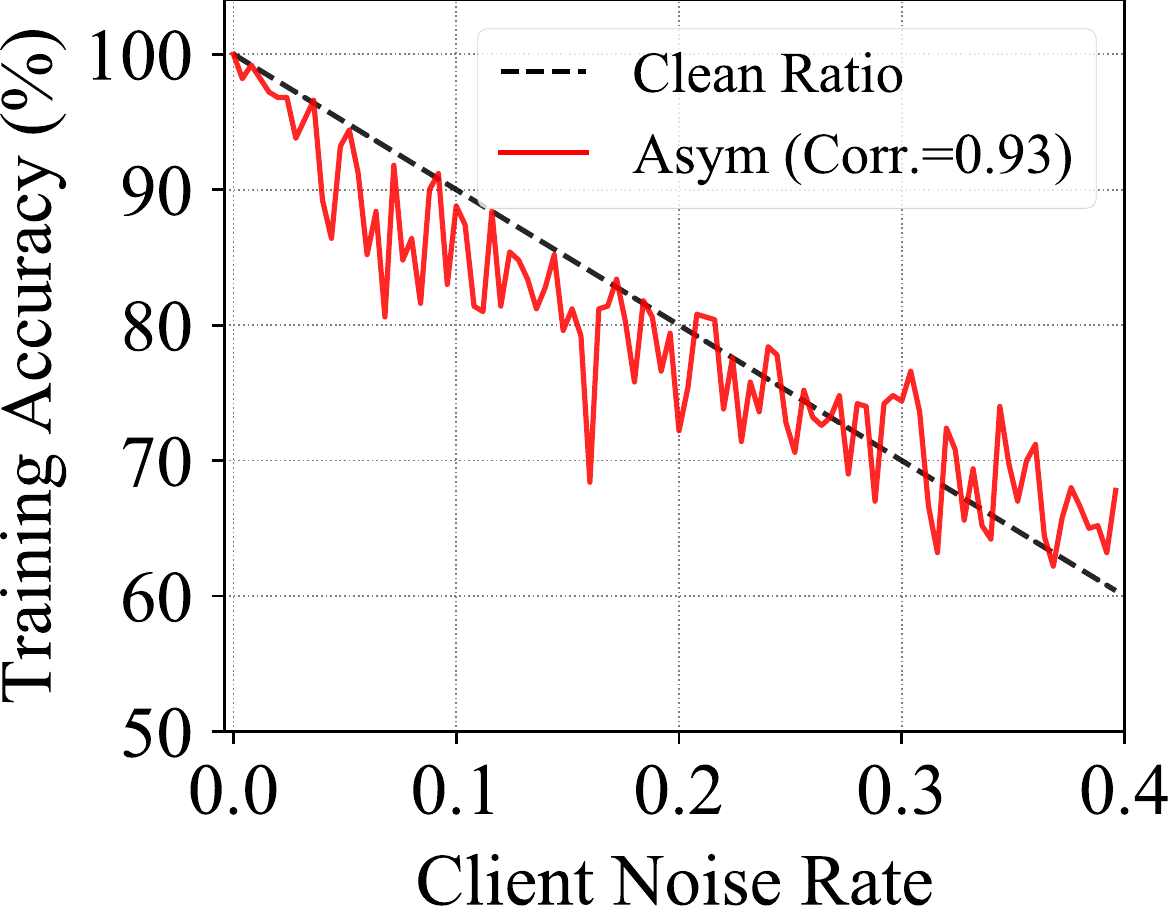}
    \label{fig:expertise-asym}
\end{subfigure}
\\
\vspace*{-0.4cm}
\hspace*{0.55cm} {\small (a) Symmetric Noise.} \hspace*{2.0cm} {\small (b) Asymmetric Noise.}
\vspace*{-0.3cm}
\caption{Correlation between the training accuracy and noise rate on CIFAR-10 with shard\,($S=2$) of symmetric and asymmetric noises $0.0$--$0.4$.}
\vspace*{-0.1cm}
\label{fig:expertise-noniid}
\end{figure}

\vspace*{-0.05cm}
\subsection{Robustness Comparison}
\label{sec:robust_comparison}

\subsubsection{Results with Synthetic Noise} 
Tables \ref{tab:cifar10} and \ref{tab:cifar100} show the test accuracy of the global model trained by eight methods for three Non-IID federated learning scenarios with four different noise settings, including symmetric, asymmetric, and mixed label noise. 
Overall, \algname{} achieves the highest test accuracy in every case.
Compared with FedAvg, \algname{}'s accuracy improves relatively more as the noise rate increases from $0.0$--$0.4$ to $0.0$--$0.8$ in symmetric noise, because the performance discrepancy over clients drastically degrades with the increase in noise rate. 
In contrast, the other robust methods show considerably poor performance compared with \algname{}, which is presumably attributed to the client-to-client performance discrepancy, despite the fact that they are incorporated with FedAvg. 
Meanwhile, the recent robust FL method, Robust FL, shows relatively poor performance in our setup since the method produces many false label corrections due to the clients with heavy label noise. Unlike, \algname{} is robust even to heavy noise because all unreliable labels are excluded from training for high safety without correction. Our method can be extended with semi-supervised learning for exploiting even unreliable examples. We leave this as future work. \looseness=-1



\subsubsection{Results with Real-world Noise} 
Table \ref{tab:webvision} displays the validation accuracy of seven different methods on a real-world noisy mini-WebVision dataset.
We report the top-1 and top-5 classification accuracy on the mini-WebVision validation set.
\algname{} maintains its performance dominance over multiple state-of-the-art robust methods for \textit{real-world} label noise as well. \algname{}\,($k=2$) improves the top-1 accuracy by up to $9.2$ compared with other robust methods. \looseness=-1

\begin{figure}
\centering
\begin{subfigure}[b]{0.45\linewidth}
\centering
\includegraphics[width=\textwidth]{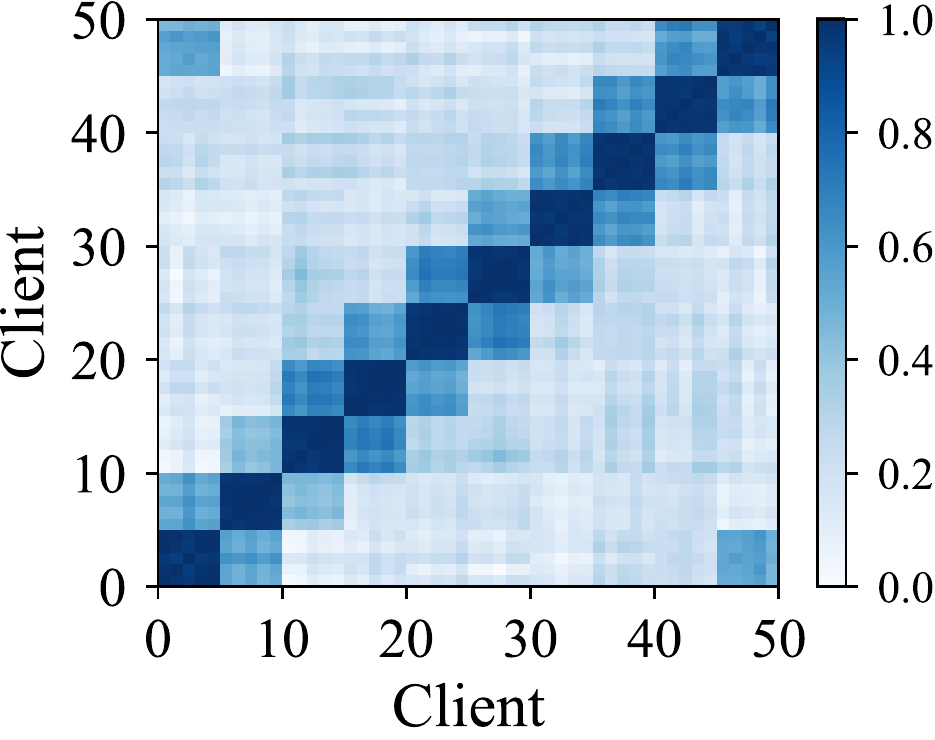}
\label{fig:sim-clean}
\vspace*{-0.5cm}
\end{subfigure}
\hfill
\begin{subfigure}[b]{0.45\linewidth}
\centering
\includegraphics[width=\textwidth]{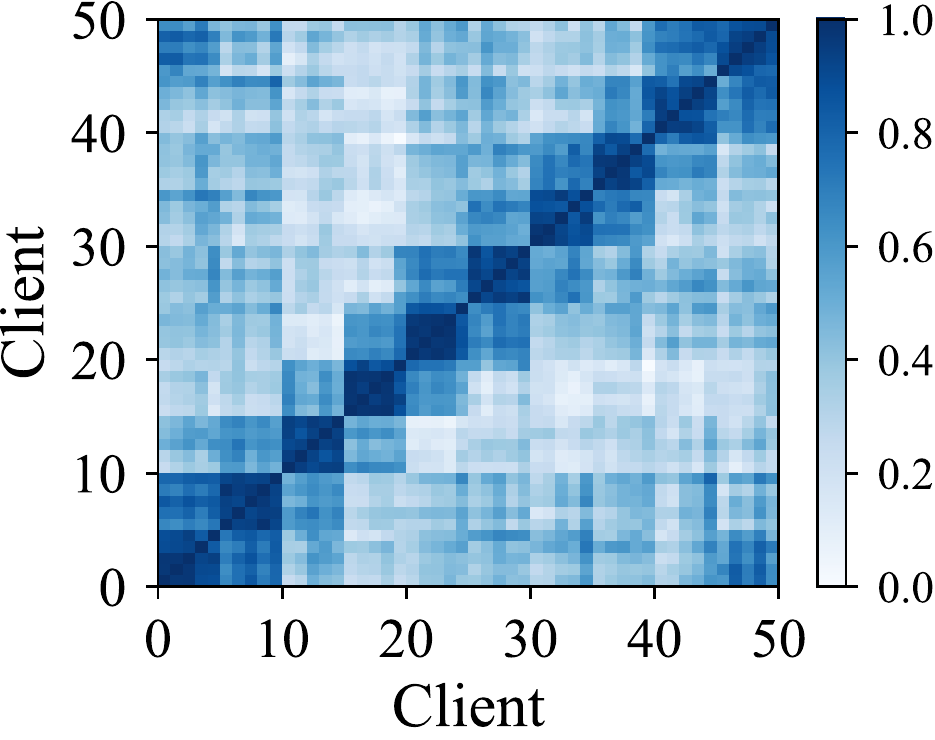}
\label{fig:sim-noisy}
\vspace*{-0.5cm}
\end{subfigure}
\\
{\small (a) w.o Label Noise.} \hspace*{2.0cm} {\small (b) w. Label Noise.}
\vspace*{-0.25cm}
\caption{Similarity matrix between the client's softmax output for a Gaussian random noise when trained on CIFAR-10 with shard\,($S=2$) without noise (a) and with symmetric noise $0.0$--$0.4$ (b).}
\vspace*{-0.3cm}
\label{fig:similarity}
\end{figure}

\subsection{In-depth Analysis on Selected Examples} 
\label{sec:selection_comparison}


All methods except FedAvg and Joint Optimization follow the pipeline of learning with sample selection. Hence, we evaluate the sample selection performance of them using the \textit{two} metrics, namely label precision\,(LP) and recall\,(LR), which respectively represent the quality and quantity of examples selected as clean\,\cite{han2018co, song2021robust}, 

{
{\color{white}{empty}}
\vspace*{-0.3cm}
\begin{equation*}
{\rm LP}=\frac{|\{(x, \tilde{y}) \in \mathcal{S}_{\sf c} : \tilde{y} = y^{*} \}|}{|\mathcal{S}_{\sf c}|}, ~ {\rm LR} = \frac{|\{(x, \tilde{y}) \in \mathcal{S}_{\sf c} : \tilde{y} = y^{*} \}|}{|\{(x, \tilde{y}) \in \mathcal{D}_{\sf c} : \tilde{y} = y^{*} \}|},
\end{equation*}
}
where $\mathcal{D}_{\sf c}$ and $\mathcal{S}_{\sf c}$ denote the training data and its selected clean set of the client ${\sf c}$, respectively. 

Figure \ref{fig:label-pr} shows the LP and LR curves averaged across all the clients per epoch on CIFAR-10 with symmetric and asymmetric label noise. The results of Robust FL are omitted due to its too low label precision and recall under the Non-IID scenario. After the warm-up period of $100$ transmission rounds, \algname{} shows the highest LP and LR with a large improvement of up to $0.12$ and $0.11$, respectively. In addition, its dominance over other robust learning methods is consistent regardless of the noise type. Therefore, these results indicate that the superior robustness of \algname{} in Tables \ref{tab:cifar10}--\ref{tab:webvision} is due to its high LP and LR. 

\subsection{In-depth Analysis on Reliability Metric}
\label{sec:in_depth_metric}
\vspace*{0.1cm}

We justify the use of two indicators for data expertise and similarity in Eqs.\,\eqref{eq:expertise} and \eqref{eq:similarity}.

\begin{itemize}[leftmargin=*]
\item \underline{Data Expertise:} Figure \ref{fig:expertise-noniid} shows the training convergence rate (training accuracy) of all clients sorted in ascending order by their noise rate. Owing to the memorization effect of DNNs, the training accuracy of the local model indeed has a strong correlation of over $0.93$ with respect to varying levels of label noise. These results confirm that the client's training accuracy represents its data expertise.
\vspace*{0.05cm}
\item \underline{Data Similarity:} Figure \ref{fig:similarity} shows similarity matrices between client's softmax outputs for a Gaussian random noise when trained on data with or without label noise. As clients with similar Ids\,(indices) have similar data distributions in our Non-IID setting, the similarity around diagonal entries is high, as can be seen from Figure \ref{fig:similarity}(a). Even with label noise shown in Figure \ref{fig:similarity}(b), it turns out that the similarity trend remains. Therefore, it can be concluded that the similarity between the softmax output is a robust metric to find clients with similar data distributions.  
\end{itemize}
%
These empirical studies provide empirical evidence for the use of our two indicators, ${\rm Exp}(\cdot)$ and ${\rm Sim}(\cdot,\cdot)$.

\subsection{Reliable Neighbors with Fine-tuning}
\label{sec:analysis_fine_tune}
\vspace*{0.05cm}

We analyze the effect of fine-tuning on data expertise and similarity used in neighbor search.
Table \ref{tab:ablation-finetune} summarizes the performance of \algname{}\,($k=2$) {with} and {without} fine-tuning, where a small $\alpha$ indicates that data similarity is considered more than data expertise.
When fine-tuning is deactivated, the benefit of retrieving neighbors with high data similarity is remarkably dominant, because lower data similarity implies a larger distribution shift between the target and neighbor models.
On the other hand, an opposite trend is observed when fine-tuning is activated. The test accuracy is high when data expertise is properly considered in conjunction with data similarity. Overall, fine-tuning does not only help mitigate the distribution shift problem but also significantly improves performance of \algname{}. 

{
\newcolumntype{L}[1]{>{\raggedright\let\newline\\\arraybackslash\hspace{0pt}}m{#1}}
\newcolumntype{X}[1]{>{\centering\let\newline\\\arraybackslash\hspace{0pt}}p{#1}}
\begin{table}[!t]
\small
\begin{tabular}{L{0.55cm} |X{1.55cm} X{1.55cm} |X{1.55cm} X{1.55cm}} \toprule
            & \multicolumn{2}{c|}{w.o. Fine-tuning}  & \multicolumn{2}{c}{w. Fine-tuning} \\
$\alpha$    & \!\!\!0.0--0.4\!\!\! & \!\!\!0.0--0.8\!\!\! & \!\!\!0.0--0.4\!\!\! & \!\!\!0.0--0.8\!\!\! \\\midrule 
0.0         & \textbf{65.03} & \textbf{58.20 }& 67.11 & 61.81 \\ 
0.2         & 63.77 & 57.24 & 67.45 & 62.69 \\ 
0.4         & 62.45 & 53.89 & 67.03 & 62.94 \\ 
0.6         & 58.74 & 51.36 & \textbf{67.62} & 62.94 \\ 
0.8         & 56.60 & 43.61 & 67.19 & \textbf{62.97} \\ 
1.0         & 56.06 & 44.87 & 66.73 & 61.61 \\\bottomrule 
\end{tabular}
\vspace*{0.1cm}
\caption{Effect of fine-tuning when trained on CIFAR-10 with symmetric noise and shard\,($S=2$) settings.}
\label{tab:ablation-finetune}
\vspace*{-0.6cm}
\end{table}
}

\vspace*{-0.1cm}
\subsection{Ablation Study on k-Reliable Neighbors}
\label{sec:aba_neighbor}
\vspace*{0.05cm}

{
\newcolumntype{L}[1]{>{\raggedright\let\newline\\\arraybackslash\hspace{0pt}}m{#1}}
\newcolumntype{X}[1]{>{\centering\let\newline\\\arraybackslash\hspace{0pt}}p{#1}}
\begin{table}[!t]
\small
\begin{tabular}{L{0.55cm} |X{1.55cm} X{1.55cm} |X{1.55cm} X{1.55cm}} \toprule
            & \multicolumn{2}{c|}{Shard ($S=2$)}  & \multicolumn{2}{c}{Dirichlet ($\beta=0.5$)} \\
$k$    & 0.0--0.4 & 0.0--0.8 & 0.0--0.4 & 0.0--0.8  \\ \midrule
1 & {67.33} & 60.33 & 79.99 & 75.92 \\
2 & 67.62 & {62.94} & {80.34} & {76.49} \\
3 & {68.10} & 62.77 & 80.31 & 76.01 \\
4 & 67.38 & 62.75 & {80.36} & {76.16} \\
5 & 67.60 & {63.10} & 80.18 & 76.00 \\
\bottomrule
\end{tabular}
\vspace*{0.1cm}
\caption{Test accuracy\,(\%) on CIFAR-10 using \algname{} with different number of reliable neighbors.}
\label{tab:ablation-neighbor}
\vspace*{-0.7cm}
\end{table}
}

{
\newcolumntype{L}[1]{>{\raggedright\let\newline\\\arraybackslash\hspace{0pt}}m{#1}}
\newcolumntype{X}[1]{>{\centering\let\newline\\\arraybackslash\hspace{0pt}}p{#1}}
\begin{table}[!t]
\small
\begin{tabular}{L{0.55cm} |X{1.55cm} X{1.55cm} |X{1.55cm} X{1.55cm}} \toprule
 &  \multicolumn{4}{c}{Shard\,($S=20$)} \\\midrule
 & \multicolumn{2}{c|}{Symmetric} & {Asym} & {Mixed} \\
$k$ & \!\!0.0--0.4\!\! & \!\!0.0--0.8\!\! & \!\!0.0--0.4\!\! & \!\!0.0--0.4\!\! \\\toprule
1 & 47.30 & 42.27 & 47.62 & 46.62 \\
2 & 47.46 & {43.07} & 47.52 & 47.17 \\
3 & 47.65 & 42.33 & {48.18} & 47.01 \\
4 & 47.66 & 42.48 & 47.55 & {47.72} \\
5 & {48.10} & 42.58 & 47.43 & 47.40 \\
\bottomrule
\end{tabular}
\vspace*{0.1cm}
\caption{Test accuracy\,(\%) of on CIFAR-100 using \algname{} with different number of reliable neighbors.}
\label{tab:cifar100-neighbor}
\vspace*{-0.7cm}
\end{table}
}

{
\newcolumntype{L}[1]{>{\raggedright\let\newline\\\arraybackslash\hspace{0pt}}m{#1}}
\newcolumntype{X}[1]{>{\centering\let\newline\\\arraybackslash\hspace{0pt}}p{#1}}
\begin{table*}[!ht]
\small
\centering
\begin{tabular}{c|cc|c|cc|c}
\toprule
 & \multicolumn{3}{c|}{Communication}   & \multicolumn{3}{c}{Computation} \\
\midrule
 & Server$\rightarrow$Client & Client$\rightarrow$Server & \textbf{Total Cost} & Forward & Backward & \textbf{Total Cost} \\\midrule
FedAvg\,\cite{mcmahan2017communication}   & $\textbf{M}$          & $\textbf{M}$            & $2\textbf{M}$   & $e\textbf{F}$  &$e\textbf{B}$ &    $e\textbf{F} + e\textbf{B}$      \\ 
Co-Teaching\,\cite{han2018co}  & $2\textbf{M}$          & $2\textbf{M}$   & $4\textbf{M}$   & $2e\textbf{F}$  &$2e\textbf{B}$ &    $2e\textbf{F} + 2e\textbf{B}$      \\ 
Joint-optm\,\cite{tanaka2018joint}   & $\textbf{M}$          & $\textbf{M}$            & $2\textbf{M}$   & $e\textbf{F}$  &$e\textbf{B}$ &    $e\textbf{F} + e\textbf{B}$      \\ 
SELFIE\,\cite{song2019selfie}  & $\textbf{M}$          & $\textbf{M}$            & $2\textbf{M}$   & $e\textbf{F}$  &$e\textbf{B}$ &    $2e\textbf{F} + e\textbf{B}$      \\ 
DivideMix\,\cite{li2020dividemix}  & $2\textbf{M}$          & $2\textbf{M}$   & $4\textbf{M}$   & $(4m+2)e\textbf{F}$  &$2e\textbf{B}$ &    $(4m+2)e\textbf{F} + 2e\textbf{B}$      \\ 
Robust FL\,\cite{yang2022robust}   & $\textbf{M}$          & $\textbf{M}$            & $2\textbf{M}$   & $e\textbf{F}$  &$e\textbf{B}$ &    $e\textbf{F} + e\textbf{B}$      \\ 
FedRN  & $(k+1)\textbf{M}$          & $\textbf{M}$   & $(k+2)\textbf{M}$   & $(e+2k+1)\textbf{F}$  &$(k+e)\textbf{B}$ &   $(e+2k+1)\textbf{F} + (k+e)\textbf{B}$      \\ 
\bottomrule
\end{tabular}%
 \vspace*{0.1cm}
 \caption{Analysis of the communication and computation costs in federated learning setting: $\textbf{M}$ is the communication cost to send the model; $\textbf{F}$ and $\textbf{B}$ are the computational costs of forward and backward propagation, respectively; $e$ is the number of local epochs for each communication round; $k$ is the number of reliable neighbors.}
\label{tab:analysis-cost}
 \vspace*{-0.7cm}
\end{table*}
}

A larger number of reliable neighbors could increase the communication overhead in federated learning. 
We, therefore, investigated the change in performance of the global model according to the the number of reliable neighbors. 
Tables \ref{tab:ablation-neighbor} and \ref{tab:cifar100-neighbor} show the classification accuracy on CIFAR-10 and CIFAR-100 respectively with varying number of neighbors. 
The results show that \algname{} with $k=1$ and $k=2$ generally shows comparable performance than $k>2$;
there is little improvement in performance with the increase in the number of reliable neighbors as long as $k>1$. In other words, using only one or two reliable neighbors is sufficient to improve the robustness, because the proposed reliability score successfully retrieves the most helpful neighbor client among all the neighbors. Therefore, in a scenario where the communication burden is significant, practitioners may reduce the communication overhead by only using a single reliable neighbor with \algname{} to handle the problem of noisy labels.

\subsection{Grid Search for Balancing Term} 
\label{sec:parameter_search}

We adjusted $\alpha$ to determine the contribution of data expertise and similarity in neighbor search. To ascertain the optimal value, we repeated the experiment on CIFAR-10 with various $\alpha$ values, as summarized in Table \ref{tab:ablation-alpha}. Because the fine-tuning technique is incorporated in \algname{}, merging data expertise and similarity provides a synergistic enhancement in general. In particular, when $\alpha = 0.6$, the performance gain is the highest on average. 
Based on these results,
we set $\alpha$ to $0.6$ for all experiments in Section \ref{sec:evaluation}.








{
\newcolumntype{L}[1]{>{\raggedright\let\newline\\\arraybackslash\hspace{0pt}}m{#1}}
\newcolumntype{X}[1]{>{\centering\let\newline\\\arraybackslash\hspace{0pt}}p{#1}}
\begin{table}[t!]
\small
\begin{tabular}{L{1.4cm} |X{0.75cm} X{0.75cm} |X{0.75cm} X{0.75cm} |X{0.75cm} X{0.75cm}} \toprule
 &  \multicolumn{6}{c}{Symmetric Noise}  \\\midrule
{\vspace*{-0.3cm}Type} &  \multicolumn{2}{c|}{Shard\,($S=2$)} &  \multicolumn{2}{c|}{Shard\,($S=5$)}  &  \multicolumn{2}{c}{Dirichlet}  \\
 &  \!\!\!\! \!\!0.0--0.4\!\!\!\! \!\! & \!\!\!\! \!\!0.0--0.8 \!\!\!\! \!\! & \!\!\!\! \!\!0.0--0.4\!\!\!\! \!\!   & \!\!\!\! \!\!0.0--0.8\!\!\!\! \!\!  & \!\!\!\! \!\!0.0--0.4\!\!\!\! \!\! &\!\!\!\!\!\!\!\!\ 0.0--0.8 \!\!\!\!\!\!\!\!\\\midrule
\!$k$-random\!\!\!\!\!   &  66.82 & 60.11 & 76.37 &  71.56 & 79.79 &  75.11 \\
\!$k$-reliable\!\!\!\!\! & \textbf{67.62} &  \textbf{62.94} &  \textbf{76.81} &  \textbf{72.85} &  \textbf{80.34} & \textbf{76.49} \\
\bottomrule
\end{tabular}
\vspace*{0.1cm}
\caption{Performance comparison with $k$-random neighbors on CIFAR-10 when $k=2$.}
\label{tab:ablation-random}
\vspace*{-0.7cm}
\end{table}
}

\subsection{Comparison with Random Neighbors}
\label{sec:random_analysis}

We analyzed the usefulness of $k$-reliable neighbors compared with $k$-random neighbors. For $k$-random neighbors, we replaced Lines 6 and 10 of Algorithm \ref{alg:fedrn} with random client selection.
%
Table \ref{tab:ablation-random} summarizes the test accuracy of \algname{} with $k$-reliable and $k$-random neighbors. 
Since our fine-tuning technique was incorporated with $k$-random neighbors, using $k$-random neighbors improves the robustness against noisy labels, but \algname{} with $k$-reliable neighbors maintains its dominance in every case.
In detail, if the neighbors with low data expertise are selected, they misclassify mislabeled examples as clean examples in sample selection. Likewise, the neighbors with significantly different data distributions harm identifying clean examples from the target client’s noisy data due to the distribution shift problem. It can be confirmed experimentally that this issue becomes more prominent when the noise rate is high. In conclusion, we should leverage the reliability of neighbors to obtain a global model robust to noisy labels.

\subsection{Detailed Cost Analysis}
\label{sec:cost_analysis}

We analyze the communication and computation cost of \algname{} compared with other {six} methods, as summarized in Table \ref{tab:analysis-cost}.
In this comparison, the communication cost is split into two parts, \enquote{server$\rightarrow$client} and \enquote{client$\rightarrow$server,} while the computation cost is split into two other parts, \enquote{forward} and \enquote{backward,} and the \enquote{total cost} aggregates all the costs for each perspective. 

{
\newcolumntype{L}[1]{>{\raggedright\let\newline\\\arraybackslash\hspace{0pt}}m{#1}}
\newcolumntype{X}[1]{>{\centering\let\newline\\\arraybackslash\hspace{0pt}}p{#1}}
\begin{table}[!t]
\small
\begin{tabular}{X{0.6cm} |X{1.0cm} X{1.0cm} |X{1.0cm} X{1.0cm} | X{1.75cm} } \toprule
            & \multicolumn{2}{c|}{Shard\,($S=2$)}  & \multicolumn{2}{c|}{Dirichlet\,($\beta=0.5$)} &   \\ \midrule
$\alpha$    & 0.0--0.4 & 0.0--0.8 & 0.0--0.4 & 0.0--0.8 &  Mean  \\ \midrule
0.0 & 67.11 & 61.81 & 80.18 & 75.94 & 71.26 $\pm$ 8.33 \\
0.2 & 67.45 & 62.69 & 80.35 & 75.95 & 71.61 $\pm$ 8.00 \\
0.4 & 67.03 & 62.94 & 79.95 & 76.27 & 71.55 $\pm$ 7.90 \\
0.6 & 67.62 & 62.94 & 80.34 & 76.49 & \textbf{71.85 $\pm$ 7.97} \\
0.8 & 67.19 & 62.97 & 80.18 & 76.28 & 71.76 $\pm$ 7.80 \\
1.0 & 66.73 & 61.61 & 80.00 & 76.07 & 71.10 $\pm$ 8.43 \\
\bottomrule
\end{tabular}
\vspace*{0.0cm}
\caption{Test accuracy\,(\%) with different $\alpha$ values.}
\label{tab:ablation-alpha}
\vspace*{-0.6cm}
\end{table}
}

\begin{itemize}[leftmargin=*]
\item \underline{Communication Cost:} \algname{} requires {3M} or {4M} communication cost in total when using one or two reliable neighbors, which is a sufficient number for general use cases. The remaining communication cost for calculating data similarity and expertise\,(i.e., $Acc({\sf c})$ and $p(\tilde{x};\Theta_{\sf c})$) is negligible. By contrast, Co-Teaching and DivideMix both require 4M communication cost in total because they maintain two models for co-training. In summary, \algname{} leads to greater performance with comparable or lower communication cost compared with other baseline methods.
\smallskip


\item \underline{Computation Cost:} We exclude the cost of fitting GMMs and searching $k$-reliable neighbors because their costs are negligible compared to those of the forward and backward steps. Along the pipeline of our local update phase, \algname{} first requires $k$F and $k$B computation cost to perform fine-tuning with $k$-reliable neighbors for one epoch. Next, to construct the clean set, $(k+1)$F communication cost is needed due to the inference with $k+1$ available models. As the constructed clean set is deterministic over all the local epochs, only $e$F+$e$B computation cost is required to train a single model for the given local epoch $e$ in each communication round. Consequently, \algname{} needs $(e+2k+1)$F + $(k+e)$B computation cost in total. 
In a typical deep learning pipeline, the cost of the backward step is relatively smaller than that of the forward step. Hence, the computation cost for the forward step is the major issue. In light of this fact, the forward cost of \algname{} is comparable to those of FedAvg, Joint-optim, and SELFIE, and less than those of Co-teaching and DivideMix, because the number of local epochs is larger than the number of reliable neighbors, $e > k$. Among the compared methods, DivideMix is the slowest method in that it performs $m$ data augmentations with two models for each local epoch, so its forward cost is $(4m+2)e$F. 
\end{itemize}

Therefore, with one or two reliable neighbors, \algname{} greatly improves the robustness of the global model without adding much communication and computation cost in the scenario of federated learning with noisy labels.
\balance

\vspace*{-0.1cm}
\section{Conclusion}
\label{sec:conclusion}

The scenario of considering both data heterogeneity and label noise is a new important research direction of practical federated learning.
In this paper, we propose a novel federated learning method called FedRN. In the local update phase, $k$-reliable neighbors with high data expertise or similarity are retrieved, and they identify clean examples collaboratively based on their reliability scores. The models are then trained on the clean set, and the weights are averaged in the model aggregation phase. 
As a result, \algname{} achieves high label precision and recall for sample selection, leading to a robust global model even in the presence of label noise. We conducted extensive experiments on three real-world or synthetic noisy datasets. The results verified that \algname{} improves the robustness against label noise consistently and significantly. Overall, the use of reliable neighbors will inspire future studies for robustness.

\section{Acknowledgement}
This work was supported by Institute of Information \& communications Technology Planning \& Evaluation (IITP) grant funded by the Korea government(MSIT) (No.2019-0-00075, Artificial Intelligence Graduate School Program(KAIST) and No. 2022-0-00871, Development of AI Autonomy and Knowledge Enhancement for AI Agent Collaboration) 
\balance

\bibliographystyle{ACM-Reference-Format}
\vfill\eject

\bibliography{0-paper.bbl}


\begin{thebibliography}{44}


\ifx \showCODEN    \undefined \def \showCODEN     #1{\unskip}     \fi
\ifx \showDOI      \undefined \def \showDOI       #1{#1}\fi
\ifx \showISBNx    \undefined \def \showISBNx     #1{\unskip}     \fi
\ifx \showISBNxiii \undefined \def \showISBNxiii  #1{\unskip}     \fi
\ifx \showISSN     \undefined \def \showISSN      #1{\unskip}     \fi
\ifx \showLCCN     \undefined \def \showLCCN      #1{\unskip}     \fi
\ifx \shownote     \undefined \def \shownote      #1{#1}          \fi
\ifx \showarticletitle \undefined \def \showarticletitle #1{#1}   \fi
\ifx \showURL      \undefined \def \showURL       {\relax}        \fi
\providecommand\bibfield[2]{#2}
\providecommand\bibinfo[2]{#2}
\providecommand\natexlab[1]{#1}
\providecommand\showeprint[2][]{arXiv:#2}

\bibitem[\protect\citeauthoryear{Acar, Zhao, Matas, Mattina, Whatmough, and
  Saligrama}{Acar et~al\mbox{.}}{2020}]%
        {acar2020federated}
\bibfield{author}{\bibinfo{person}{Durmus Alp~Emre Acar}, \bibinfo{person}{Yue
  Zhao}, \bibinfo{person}{Ramon Matas}, \bibinfo{person}{Matthew Mattina},
  \bibinfo{person}{Paul Whatmough}, {and} \bibinfo{person}{Venkatesh
  Saligrama}.} \bibinfo{year}{2020}\natexlab{}.
\newblock \showarticletitle{Federated learning based on dynamic
  regularization}. In \bibinfo{booktitle}{\emph{ICLR}}.
\newblock


\bibitem[\protect\citeauthoryear{Arazo, Ortego, Albert, O’Connor, and
  McGuinness}{Arazo et~al\mbox{.}}{2019}]%
        {arazo2019unsupervised}
\bibfield{author}{\bibinfo{person}{Eric Arazo}, \bibinfo{person}{Diego Ortego},
  \bibinfo{person}{Paul Albert}, \bibinfo{person}{Noel O’Connor}, {and}
  \bibinfo{person}{Kevin McGuinness}.} \bibinfo{year}{2019}\natexlab{}.
\newblock \showarticletitle{Unsupervised label noise modeling and loss
  correction}. In \bibinfo{booktitle}{\emph{ICML}}.
\newblock


\bibitem[\protect\citeauthoryear{Arpit, Jastrz{{e}}bski, Ballas, Krueger,
  Bengio, Kanwal, Maharaj, Fischer, Courville, Bengio, et~al\mbox{.}}{Arpit
  et~al\mbox{.}}{2017}]%
        {arpit2017closer}
\bibfield{author}{\bibinfo{person}{Devansh Arpit},
  \bibinfo{person}{Stanis{\l}aw Jastrz{{e}}bski}, \bibinfo{person}{Nicolas
  Ballas}, \bibinfo{person}{David Krueger}, \bibinfo{person}{Emmanuel Bengio},
  \bibinfo{person}{Maxinder~S Kanwal}, \bibinfo{person}{Tegan Maharaj},
  \bibinfo{person}{Asja Fischer}, \bibinfo{person}{Aaron Courville},
  \bibinfo{person}{Yoshua Bengio}, {et~al\mbox{.}}}
  \bibinfo{year}{2017}\natexlab{}.
\newblock \showarticletitle{A closer look at memorization in deep networks}. In
  \bibinfo{booktitle}{\emph{ICML}}.
\newblock


\bibitem[\protect\citeauthoryear{Berthelot, Carlini, Goodfellow, Papernot,
  Oliver, and Raffel}{Berthelot et~al\mbox{.}}{2019}]%
        {berthelot2019mixmatch}
\bibfield{author}{\bibinfo{person}{David Berthelot}, \bibinfo{person}{Nicholas
  Carlini}, \bibinfo{person}{Ian Goodfellow}, \bibinfo{person}{Nicolas
  Papernot}, \bibinfo{person}{Avital Oliver}, {and} \bibinfo{person}{Colin
  Raffel}.} \bibinfo{year}{2019}\natexlab{}.
\newblock \showarticletitle{Mix{M}atch: {A} holistic approach to
  semi-supervised learning}. In \bibinfo{booktitle}{\emph{NeurIPS}}.
\newblock


\bibitem[\protect\citeauthoryear{Bonawitz, Eichner, Grieskamp, Huba, Ingerman,
  Ivanov, Kiddon, Kone{\v{c}}n{\`y}, Mazzocchi, McMahan,
  et~al\mbox{.}}{Bonawitz et~al\mbox{.}}{2019}]%
        {bonawitz2019towards}
\bibfield{author}{\bibinfo{person}{Keith Bonawitz}, \bibinfo{person}{Hubert
  Eichner}, \bibinfo{person}{Wolfgang Grieskamp}, \bibinfo{person}{Dzmitry
  Huba}, \bibinfo{person}{Alex Ingerman}, \bibinfo{person}{Vladimir Ivanov},
  \bibinfo{person}{Chloe Kiddon}, \bibinfo{person}{Jakub Kone{\v{c}}n{\`y}},
  \bibinfo{person}{Stefano Mazzocchi}, \bibinfo{person}{H~Brendan McMahan},
  {et~al\mbox{.}}} \bibinfo{year}{2019}\natexlab{}.
\newblock \showarticletitle{Towards federated learning at scale: {S}ystem
  design}. In \bibinfo{booktitle}{\emph{SysML}}.
\newblock


\bibitem[\protect\citeauthoryear{Briggs, Fan, and Andras}{Briggs
  et~al\mbox{.}}{2020}]%
        {briggs2020federated}
\bibfield{author}{\bibinfo{person}{Christopher Briggs}, \bibinfo{person}{Zhong
  Fan}, {and} \bibinfo{person}{Peter Andras}.} \bibinfo{year}{2020}\natexlab{}.
\newblock \showarticletitle{Federated learning with hierarchical clustering of
  local updates to improve training on non-IID data}. In
  \bibinfo{booktitle}{\emph{IJCNN}}.
\newblock


\bibitem[\protect\citeauthoryear{Chen, Ding, Tramel, Wu, Sahu, Avestimehr, and
  Zhang}{Chen et~al\mbox{.}}{2022}]%
        {chen2022self}
\bibfield{author}{\bibinfo{person}{Huili Chen}, \bibinfo{person}{Jie Ding},
  \bibinfo{person}{Eric Tramel}, \bibinfo{person}{Shuang Wu},
  \bibinfo{person}{Anit~Kumar Sahu}, \bibinfo{person}{Salman Avestimehr}, {and}
  \bibinfo{person}{Tao Zhang}.} \bibinfo{year}{2022}\natexlab{}.
\newblock \showarticletitle{Self-Aware Personalized Federated Learning}.
\newblock \bibinfo{journal}{\emph{arXiv preprint arXiv:2204.08069}}
  (\bibinfo{year}{2022}).
\newblock


\bibitem[\protect\citeauthoryear{Chen and Chao}{Chen and Chao}{2021}]%
        {chen2020fedbe}
\bibfield{author}{\bibinfo{person}{Hong-You Chen} {and}
  \bibinfo{person}{Wei-Lun Chao}.} \bibinfo{year}{2021}\natexlab{}.
\newblock \showarticletitle{{FEDBE}: {M}aking bayesian model ensemble
  applicable to federated learning}. In \bibinfo{booktitle}{\emph{ICLR}}.
\newblock


\bibitem[\protect\citeauthoryear{Chen, Zhang, Guo, Fan, and Cheng}{Chen
  et~al\mbox{.}}{2021}]%
        {chen2021fedmatch}
\bibfield{author}{\bibinfo{person}{Jiangui Chen}, \bibinfo{person}{Ruqing
  Zhang}, \bibinfo{person}{Jiafeng Guo}, \bibinfo{person}{Yixing Fan}, {and}
  \bibinfo{person}{Xueqi Cheng}.} \bibinfo{year}{2021}\natexlab{}.
\newblock \showarticletitle{FedMatch: Federated Learning Over Heterogeneous
  Question Answering Data}. In \bibinfo{booktitle}{\emph{CIKM}}.
\newblock


\bibitem[\protect\citeauthoryear{Chen, Liao, Chen, and Zhang}{Chen
  et~al\mbox{.}}{2019}]%
        {chen2019understanding}
\bibfield{author}{\bibinfo{person}{Pengfei Chen}, \bibinfo{person}{Ben~Ben
  Liao}, \bibinfo{person}{Guangyong Chen}, {and} \bibinfo{person}{Shengyu
  Zhang}.} \bibinfo{year}{2019}\natexlab{}.
\newblock \showarticletitle{Understanding and utilizing deep neural networks
  trained with noisy labels}. In \bibinfo{booktitle}{\emph{ICML}}.
\newblock


\bibitem[\protect\citeauthoryear{Chen, Yang, Qin, Yu, Chen, and Shen}{Chen
  et~al\mbox{.}}{2020}]%
        {chen2020focus}
\bibfield{author}{\bibinfo{person}{Yiqiang Chen}, \bibinfo{person}{Xiaodong
  Yang}, \bibinfo{person}{Xin Qin}, \bibinfo{person}{Han Yu},
  \bibinfo{person}{Biao Chen}, {and} \bibinfo{person}{Zhiqi Shen}.}
  \bibinfo{year}{2020}\natexlab{}.
\newblock \showarticletitle{Focus: Dealing with label quality disparity in
  federated learning}. In \bibinfo{booktitle}{\emph{IJCAIW}}.
\newblock


\bibitem[\protect\citeauthoryear{Collins, Hassani, Mokhtari, and
  Shakkottai}{Collins et~al\mbox{.}}{2021}]%
        {collins2021exploiting}
\bibfield{author}{\bibinfo{person}{Liam Collins}, \bibinfo{person}{Hamed
  Hassani}, \bibinfo{person}{Aryan Mokhtari}, {and} \bibinfo{person}{Sanjay
  Shakkottai}.} \bibinfo{year}{2021}\natexlab{}.
\newblock \showarticletitle{Exploiting Shared Representations for Personalized
  Federated Learning}. In \bibinfo{booktitle}{\emph{ICML}}.
\newblock


\bibitem[\protect\citeauthoryear{Duan, Liu, Chen, Tan, Ren, Qiao, and
  Liang}{Duan et~al\mbox{.}}{2019}]%
        {duan2019astraea}
\bibfield{author}{\bibinfo{person}{Moming Duan}, \bibinfo{person}{Duo Liu},
  \bibinfo{person}{Xianzhang Chen}, \bibinfo{person}{Yujuan Tan},
  \bibinfo{person}{Jinting Ren}, \bibinfo{person}{Lei Qiao}, {and}
  \bibinfo{person}{Liang Liang}.} \bibinfo{year}{2019}\natexlab{}.
\newblock \showarticletitle{Astraea: Self-balancing federated learning for
  improving classification accuracy of mobile deep learning applications}. In
  \bibinfo{booktitle}{\emph{ICCD}}.
\newblock


\bibitem[\protect\citeauthoryear{Han, Yao, Yu, Niu, Xu, Hu, Tsang, and
  Sugiyama}{Han et~al\mbox{.}}{2018}]%
        {han2018co}
\bibfield{author}{\bibinfo{person}{Bo Han}, \bibinfo{person}{Quanming Yao},
  \bibinfo{person}{Xingrui Yu}, \bibinfo{person}{Gang Niu},
  \bibinfo{person}{Miao Xu}, \bibinfo{person}{Weihua Hu}, \bibinfo{person}{Ivor
  Tsang}, {and} \bibinfo{person}{Masashi Sugiyama}.}
  \bibinfo{year}{2018}\natexlab{}.
\newblock \showarticletitle{Co-teaching: {R}obust training of deep neural
  networks with extremely noisy labels}. In
  \bibinfo{booktitle}{\emph{NeurIPS}}.
\newblock


\bibitem[\protect\citeauthoryear{Huang, Zhang, and Zhang}{Huang
  et~al\mbox{.}}{2020}]%
        {huang2020self}
\bibfield{author}{\bibinfo{person}{Lang Huang}, \bibinfo{person}{Chao Zhang},
  {and} \bibinfo{person}{Hongyang Zhang}.} \bibinfo{year}{2020}\natexlab{}.
\newblock \showarticletitle{Self-adaptive training: beyond empirical risk
  minimization}. In \bibinfo{booktitle}{\emph{NeurIPS}}.
\newblock


\bibitem[\protect\citeauthoryear{Jeong, Yoon, Yang, and Hwang}{Jeong
  et~al\mbox{.}}{2021}]%
        {jeong2020federated}
\bibfield{author}{\bibinfo{person}{Wonyong Jeong}, \bibinfo{person}{Jaehong
  Yoon}, \bibinfo{person}{Eunho Yang}, {and} \bibinfo{person}{Sung~Ju Hwang}.}
  \bibinfo{year}{2021}\natexlab{}.
\newblock \showarticletitle{Federated semi-supervised learning with
  inter-client consistency}. In \bibinfo{booktitle}{\emph{ICLR}}.
\newblock


\bibitem[\protect\citeauthoryear{Jiang, Song, Tong, Wu, Zhao, Xu, and
  Yang}{Jiang et~al\mbox{.}}{[n.\,d.]}]%
        {jiang2019federated}
\bibfield{author}{\bibinfo{person}{Di Jiang}, \bibinfo{person}{Yuanfeng Song},
  \bibinfo{person}{Yongxin Tong}, \bibinfo{person}{Xueyang Wu},
  \bibinfo{person}{Weiwei Zhao}, \bibinfo{person}{Qian Xu}, {and}
  \bibinfo{person}{Qiang Yang}.} \bibinfo{year}{[n.\,d.]}\natexlab{}.
\newblock \showarticletitle{Federated topic modeling}. In
  \bibinfo{booktitle}{\emph{CIKM}}.
\newblock


\bibitem[\protect\citeauthoryear{Jiang, Zhou, Leung, Li, and Fei-Fei}{Jiang
  et~al\mbox{.}}{2018}]%
        {jiang2017mentornet}
\bibfield{author}{\bibinfo{person}{Lu Jiang}, \bibinfo{person}{Zhengyuan Zhou},
  \bibinfo{person}{Thomas Leung}, \bibinfo{person}{Li-Jia Li}, {and}
  \bibinfo{person}{Li Fei-Fei}.} \bibinfo{year}{2018}\natexlab{}.
\newblock \showarticletitle{Mentor{N}et: Learning data-driven curriculum for
  very deep neural networks on corrupted labels}. In
  \bibinfo{booktitle}{\emph{ICML}}.
\newblock


\bibitem[\protect\citeauthoryear{Lee, Roh, Song, and Whang}{Lee
  et~al\mbox{.}}{2021}]%
        {lee2021machine}
\bibfield{author}{\bibinfo{person}{Jae-Gil Lee}, \bibinfo{person}{Yuji Roh},
  \bibinfo{person}{Hwanjun Song}, {and} \bibinfo{person}{Steven~Euijong
  Whang}.} \bibinfo{year}{2021}\natexlab{}.
\newblock \showarticletitle{Machine Learning Robustness, Fairness, and their
  Convergence}. In \bibinfo{booktitle}{\emph{KDD}}.
\newblock


\bibitem[\protect\citeauthoryear{Li, Socher, and Hoi}{Li et~al\mbox{.}}{2020}]%
        {li2020dividemix}
\bibfield{author}{\bibinfo{person}{Junnan Li}, \bibinfo{person}{Richard
  Socher}, {and} \bibinfo{person}{Steven~CH Hoi}.}
  \bibinfo{year}{2020}\natexlab{}.
\newblock \showarticletitle{Divide{M}ix: Learning with noisy labels as
  semi-supervised learning}. In \bibinfo{booktitle}{\emph{ICLR}}.
\newblock


\bibitem[\protect\citeauthoryear{Li, Sahu, Zaheer, Sanjabi, Talwalkar, and
  Smith}{Li et~al\mbox{.}}{2018}]%
        {li2018federated}
\bibfield{author}{\bibinfo{person}{Tian Li}, \bibinfo{person}{Anit~Kumar Sahu},
  \bibinfo{person}{Manzil Zaheer}, \bibinfo{person}{Maziar Sanjabi},
  \bibinfo{person}{Ameet Talwalkar}, {and} \bibinfo{person}{Virginia Smith}.}
  \bibinfo{year}{2018}\natexlab{}.
\newblock \showarticletitle{Federated optimization in heterogeneous networks}.
\newblock \bibinfo{journal}{\emph{arXiv preprint arXiv:1812.06127}}
  (\bibinfo{year}{2018}).
\newblock


\bibitem[\protect\citeauthoryear{Luo, Yang, Ye, Guo, and Zhao}{Luo
  et~al\mbox{.}}{[n.\,d.]}]%
        {luo2021fedskel}
\bibfield{author}{\bibinfo{person}{Junyu Luo}, \bibinfo{person}{Jianlei Yang},
  \bibinfo{person}{Xucheng Ye}, \bibinfo{person}{Xin Guo}, {and}
  \bibinfo{person}{Weisheng Zhao}.} \bibinfo{year}{[n.\,d.]}\natexlab{}.
\newblock \showarticletitle{FedSkel: Efficient Federated Learning on
  Heterogeneous Systems with Skeleton Gradients Update}. In
  \bibinfo{booktitle}{\emph{CIKM}}.
\newblock


\bibitem[\protect\citeauthoryear{Malekijoo, Fadaeieslam, Malekijou,
  Homayounfar, Alizadeh-Shabdiz, and Rawassizadeh}{Malekijoo
  et~al\mbox{.}}{2021}]%
        {malekijoo2021fedzip}
\bibfield{author}{\bibinfo{person}{Amirhossein Malekijoo},
  \bibinfo{person}{Mohammad~Javad Fadaeieslam}, \bibinfo{person}{Hanieh
  Malekijou}, \bibinfo{person}{Morteza Homayounfar}, \bibinfo{person}{Farshid
  Alizadeh-Shabdiz}, {and} \bibinfo{person}{Reza Rawassizadeh}.}
  \bibinfo{year}{2021}\natexlab{}.
\newblock \showarticletitle{{FEDZIP}: {A} Compression Framework for
  Communication-Efficient Federated Learning}.
\newblock \bibinfo{journal}{\emph{arXiv preprint arXiv:2102.01593}}
  (\bibinfo{year}{2021}).
\newblock


\bibitem[\protect\citeauthoryear{McMahan, Moore, Ramage, Hampson, and
  y~Arcas}{McMahan et~al\mbox{.}}{2017}]%
        {mcmahan2017communication}
\bibfield{author}{\bibinfo{person}{Brendan McMahan}, \bibinfo{person}{Eider
  Moore}, \bibinfo{person}{Daniel Ramage}, \bibinfo{person}{Seth Hampson},
  {and} \bibinfo{person}{Blaise~Aguera y Arcas}.}
  \bibinfo{year}{2017}\natexlab{}.
\newblock \showarticletitle{Communication-efficient learning of deep networks
  from decentralized data}. In \bibinfo{booktitle}{\emph{AISTATS}}.
\newblock


\bibitem[\protect\citeauthoryear{Muhammad, Wang, O'Reilly-Morgan, Tragos,
  Smyth, Hurley, Geraci, and Lawlor}{Muhammad et~al\mbox{.}}{2020}]%
        {muhammad2020fedfast}
\bibfield{author}{\bibinfo{person}{Khalil Muhammad}, \bibinfo{person}{Qinqin
  Wang}, \bibinfo{person}{Diarmuid O'Reilly-Morgan}, \bibinfo{person}{Elias
  Tragos}, \bibinfo{person}{Barry Smyth}, \bibinfo{person}{Neil Hurley},
  \bibinfo{person}{James Geraci}, {and} \bibinfo{person}{Aonghus Lawlor}.}
  \bibinfo{year}{2020}\natexlab{}.
\newblock \showarticletitle{Fedfast: Going beyond average for faster training
  of federated recommender systems}. In \bibinfo{booktitle}{\emph{KDD}}.
\newblock


\bibitem[\protect\citeauthoryear{Patrini, Rozza, Krishna~Menon, Nock, and
  Qu}{Patrini et~al\mbox{.}}{2017}]%
        {patrini2017making}
\bibfield{author}{\bibinfo{person}{Giorgio Patrini},
  \bibinfo{person}{Alessandro Rozza}, \bibinfo{person}{Aditya Krishna~Menon},
  \bibinfo{person}{Richard Nock}, {and} \bibinfo{person}{Lizhen Qu}.}
  \bibinfo{year}{2017}\natexlab{}.
\newblock \showarticletitle{Making deep neural networks robust to label noise:
  {A} loss correction approach}. In \bibinfo{booktitle}{\emph{CVPR}}.
\newblock


\bibitem[\protect\citeauthoryear{Song, Kim, and Lee}{Song
  et~al\mbox{.}}{2019}]%
        {song2019selfie}
\bibfield{author}{\bibinfo{person}{Hwanjun Song}, \bibinfo{person}{Minseok
  Kim}, {and} \bibinfo{person}{Jae-Gil Lee}.} \bibinfo{year}{2019}\natexlab{}.
\newblock \showarticletitle{{SELFIE}: Refurbishing unclean samples for robust
  deep learning}. In \bibinfo{booktitle}{\emph{ICML}}.
\newblock


\bibitem[\protect\citeauthoryear{Song, Kim, Park, Shin, and Lee}{Song
  et~al\mbox{.}}{2021}]%
        {song2021robust}
\bibfield{author}{\bibinfo{person}{Hwanjun Song}, \bibinfo{person}{Minseok
  Kim}, \bibinfo{person}{Dongmin Park}, \bibinfo{person}{Yooju Shin}, {and}
  \bibinfo{person}{Jae-Gil Lee}.} \bibinfo{year}{2021}\natexlab{}.
\newblock \showarticletitle{Robust Learning by Self-Transition for Handling
  Noisy Labels}. In \bibinfo{booktitle}{\emph{KDD}}.
\newblock


\bibitem[\protect\citeauthoryear{Song, Kim, Park, Shin, and Lee}{Song
  et~al\mbox{.}}{2022}]%
        {song2020learning}
\bibfield{author}{\bibinfo{person}{Hwanjun Song}, \bibinfo{person}{Minseok
  Kim}, \bibinfo{person}{Dongmin Park}, \bibinfo{person}{Yooju Shin}, {and}
  \bibinfo{person}{Jae-Gil Lee}.} \bibinfo{year}{2022}\natexlab{}.
\newblock \showarticletitle{Learning from noisy labels with deep neural
  networks: A survey}.
\newblock \bibinfo{journal}{\emph{IEEE Transactions on Neural Networks and
  Learning Systems}} (\bibinfo{year}{2022}).
\newblock


\bibitem[\protect\citeauthoryear{Tanaka, Ikami, Yamasaki, and Aizawa}{Tanaka
  et~al\mbox{.}}{2018}]%
        {tanaka2018joint}
\bibfield{author}{\bibinfo{person}{Daiki Tanaka}, \bibinfo{person}{Daiki
  Ikami}, \bibinfo{person}{Toshihiko Yamasaki}, {and} \bibinfo{person}{Kiyoharu
  Aizawa}.} \bibinfo{year}{2018}\natexlab{}.
\newblock \showarticletitle{Joint optimization framework for learning with
  noisy labels}. In \bibinfo{booktitle}{\emph{CVPR}}.
\newblock


\bibitem[\protect\citeauthoryear{Tuor, Wang, Ko, Liu, and Leung}{Tuor
  et~al\mbox{.}}{2020}]%
        {tuor2021overcoming}
\bibfield{author}{\bibinfo{person}{Tiffany Tuor}, \bibinfo{person}{Shiqiang
  Wang}, \bibinfo{person}{Bong~Jun Ko}, \bibinfo{person}{Changchang Liu}, {and}
  \bibinfo{person}{Kin~K Leung}.} \bibinfo{year}{2020}\natexlab{}.
\newblock \showarticletitle{Overcoming noisy and irrelevant data in federated
  learning}. In \bibinfo{booktitle}{\emph{ICPR}}.
\newblock


\bibitem[\protect\citeauthoryear{Xia, Liu, Wang, Han, Gong, Niu, and
  Sugiyama}{Xia et~al\mbox{.}}{2019}]%
        {xia2019anchor}
\bibfield{author}{\bibinfo{person}{Xiaobo Xia}, \bibinfo{person}{Tongliang
  Liu}, \bibinfo{person}{Nannan Wang}, \bibinfo{person}{Bo Han},
  \bibinfo{person}{Chen Gong}, \bibinfo{person}{Gang Niu}, {and}
  \bibinfo{person}{Masashi Sugiyama}.} \bibinfo{year}{2019}\natexlab{}.
\newblock \showarticletitle{Are anchor points really indispensable in
  label-noise learning?}. In \bibinfo{booktitle}{\emph{NeurIPS}}.
\newblock


\bibitem[\protect\citeauthoryear{Xu, Chen, Quek, and Chong}{Xu
  et~al\mbox{.}}{2022}]%
        {xu2022fedcorr}
\bibfield{author}{\bibinfo{person}{Jingyi Xu}, \bibinfo{person}{Zihan Chen},
  \bibinfo{person}{Tony~QS Quek}, {and} \bibinfo{person}{Kai Fong~Ernest
  Chong}.} \bibinfo{year}{2022}\natexlab{}.
\newblock \showarticletitle{FedCorr: Multi-Stage Federated Learning for Label
  Noise Correction}.
\newblock \bibinfo{journal}{\emph{arXiv preprint arXiv:2204.04677}}
  (\bibinfo{year}{2022}).
\newblock


\bibitem[\protect\citeauthoryear{Yang, Wang, Xu, Chen, Bian, Liu, and Liu}{Yang
  et~al\mbox{.}}{2021a}]%
        {yang2021characterizing}
\bibfield{author}{\bibinfo{person}{Chengxu Yang}, \bibinfo{person}{Qipeng
  Wang}, \bibinfo{person}{Mengwei Xu}, \bibinfo{person}{Zhenpeng Chen},
  \bibinfo{person}{Kaigui Bian}, \bibinfo{person}{Yunxin Liu}, {and}
  \bibinfo{person}{Xuanzhe Liu}.} \bibinfo{year}{2021}\natexlab{a}.
\newblock \showarticletitle{Characterizing Impacts of Heterogeneity in
  Federated Learning upon Large-Scale Smartphone Data}. In
  \bibinfo{booktitle}{\emph{TheWebConf}}.
\newblock


\bibitem[\protect\citeauthoryear{Yang, Zhang, Hao, Spell, and Carin}{Yang
  et~al\mbox{.}}{2021b}]%
        {yang2021flop}
\bibfield{author}{\bibinfo{person}{Qian Yang}, \bibinfo{person}{Jianyi Zhang},
  \bibinfo{person}{Weituo Hao}, \bibinfo{person}{Gregory~P Spell}, {and}
  \bibinfo{person}{Lawrence Carin}.} \bibinfo{year}{2021}\natexlab{b}.
\newblock \showarticletitle{Flop: Federated learning on medical datasets using
  partial networks}. In \bibinfo{booktitle}{\emph{KDD}}.
\newblock


\bibitem[\protect\citeauthoryear{Yang, Park, Byun, and Kim}{Yang
  et~al\mbox{.}}{2022}]%
        {yang2022robust}
\bibfield{author}{\bibinfo{person}{Seunghan Yang}, \bibinfo{person}{Hyoungseob
  Park}, \bibinfo{person}{Junyoung Byun}, {and} \bibinfo{person}{Changick
  Kim}.} \bibinfo{year}{2022}\natexlab{}.
\newblock \showarticletitle{Robust federated learning with noisy labels}.
\newblock \bibinfo{journal}{\emph{IEEE Intelligent Systems}}
  (\bibinfo{year}{2022}).
\newblock


\bibitem[\protect\citeauthoryear{Yang, Andrew, Eichner, Sun, Li, Kong, Ramage,
  and Beaufays}{Yang et~al\mbox{.}}{2018}]%
        {yang2018applied}
\bibfield{author}{\bibinfo{person}{Timothy Yang}, \bibinfo{person}{Galen
  Andrew}, \bibinfo{person}{Hubert Eichner}, \bibinfo{person}{Haicheng Sun},
  \bibinfo{person}{Wei Li}, \bibinfo{person}{Nicholas Kong},
  \bibinfo{person}{Daniel Ramage}, {and} \bibinfo{person}{Fran{\c{c}}oise
  Beaufays}.} \bibinfo{year}{2018}\natexlab{}.
\newblock \showarticletitle{Applied federated learning: Improving google
  keyboard query suggestions}.
\newblock \bibinfo{journal}{\emph{arXiv preprint arXiv:1812.02903}}
  (\bibinfo{year}{2018}).
\newblock


\bibitem[\protect\citeauthoryear{Yu, Han, Yao, Niu, Tsang, and Sugiyama}{Yu
  et~al\mbox{.}}{2019}]%
        {yu2019does}
\bibfield{author}{\bibinfo{person}{Xingrui Yu}, \bibinfo{person}{Bo Han},
  \bibinfo{person}{Jiangchao Yao}, \bibinfo{person}{Gang Niu},
  \bibinfo{person}{Ivor Tsang}, {and} \bibinfo{person}{Masashi Sugiyama}.}
  \bibinfo{year}{2019}\natexlab{}.
\newblock \showarticletitle{How does disagreement help generalization against
  label corruption?}. In \bibinfo{booktitle}{\emph{ICML}}.
\newblock


\bibitem[\protect\citeauthoryear{Zeng, Yang, Chen, Yu, and Zhang}{Zeng
  et~al\mbox{.}}{2022}]%
        {zeng2022clc}
\bibfield{author}{\bibinfo{person}{Bixiao Zeng}, \bibinfo{person}{Xiaodong
  Yang}, \bibinfo{person}{Yiqiang Chen}, \bibinfo{person}{Hanchao Yu}, {and}
  \bibinfo{person}{Yingwei Zhang}.} \bibinfo{year}{2022}\natexlab{}.
\newblock \showarticletitle{CLC: A Consensus-based Label Correction Approach in
  Federated Learning}.
\newblock \bibinfo{journal}{\emph{ACM Transactions on Intelligent Systems and
  Technology}} (\bibinfo{year}{2022}).
\newblock


\bibitem[\protect\citeauthoryear{Zhang, Bengio, Hardt, Recht, and
  Vinyals}{Zhang et~al\mbox{.}}{2017}]%
        {zhang2016understanding}
\bibfield{author}{\bibinfo{person}{Chiyuan Zhang}, \bibinfo{person}{Samy
  Bengio}, \bibinfo{person}{Moritz Hardt}, \bibinfo{person}{Benjamin Recht},
  {and} \bibinfo{person}{Oriol Vinyals}.} \bibinfo{year}{2017}\natexlab{}.
\newblock \showarticletitle{Understanding deep learning requires rethinking
  generalization}. In \bibinfo{booktitle}{\emph{ICLR}}.
\newblock


\bibitem[\protect\citeauthoryear{Zhang, Gu, Dang, Deng, and Huang}{Zhang
  et~al\mbox{.}}{[n.\,d.]}]%
        {zhang2021desirable}
\bibfield{author}{\bibinfo{person}{Qingsong Zhang}, \bibinfo{person}{Bin Gu},
  \bibinfo{person}{Zhiyuan Dang}, \bibinfo{person}{Cheng Deng}, {and}
  \bibinfo{person}{Heng Huang}.} \bibinfo{year}{[n.\,d.]}\natexlab{}.
\newblock \showarticletitle{Desirable Companion for Vertical Federated
  Learning: New Zeroth-Order Gradient Based Algorithm}. In
  \bibinfo{booktitle}{\emph{CIKM}}.
\newblock


\bibitem[\protect\citeauthoryear{Zhao, Li, Lai, Suda, Civin, and Chandra}{Zhao
  et~al\mbox{.}}{2018}]%
        {zhao2018federated}
\bibfield{author}{\bibinfo{person}{Yue Zhao}, \bibinfo{person}{Meng Li},
  \bibinfo{person}{Liangzhen Lai}, \bibinfo{person}{Naveen Suda},
  \bibinfo{person}{Damon Civin}, {and} \bibinfo{person}{Vikas Chandra}.}
  \bibinfo{year}{2018}\natexlab{}.
\newblock \showarticletitle{Federated learning with non-iid data}.
\newblock \bibinfo{journal}{\emph{arXiv preprint arXiv:1806.00582}}
  (\bibinfo{year}{2018}).
\newblock


\bibitem[\protect\citeauthoryear{Zheng, Awadallah, and Dumais}{Zheng
  et~al\mbox{.}}{2021}]%
        {zheng2021meta}
\bibfield{author}{\bibinfo{person}{Guoqing Zheng},
  \bibinfo{person}{Ahmed~Hassan Awadallah}, {and} \bibinfo{person}{Susan
  Dumais}.} \bibinfo{year}{2021}\natexlab{}.
\newblock \showarticletitle{Meta label correction for noisy label learning}. In
  \bibinfo{booktitle}{\emph{AAAI}}.
\newblock


\bibitem[\protect\citeauthoryear{Zhu, Xu, Liu, and Jin}{Zhu
  et~al\mbox{.}}{2021}]%
        {zhu2021federated}
\bibfield{author}{\bibinfo{person}{Hangyu Zhu}, \bibinfo{person}{Jinjin Xu},
  \bibinfo{person}{Shiqing Liu}, {and} \bibinfo{person}{Yaochu Jin}.}
  \bibinfo{year}{2021}\natexlab{}.
\newblock \showarticletitle{Federated Learning on Non-IID Data: A Survey}.
\newblock \bibinfo{journal}{\emph{arXiv preprint arXiv:2106.06843}}
  (\bibinfo{year}{2021}).
\newblock


\end{thebibliography}


\end{document}